\documentclass[10pt,twocolumn,letterpaper]{article}

\usepackage[pagenumbers]{wacv} 

\usepackage{tikz}
\usetikzlibrary{arrows.meta, positioning, calc, backgrounds, decorations.pathreplacing, fit}

\usepackage{amsmath}          
\usepackage{amssymb}          
\usepackage{amsfonts}         
\usepackage{bm}               
\usepackage{mathtools}        

\usepackage{pgfplots}
\pgfplotsset{compat=1.18}

\usepackage[T1]{fontenc}      
\usepackage[utf8]{inputenc}   
\usepackage{microtype}        

\usepackage{graphicx}         
\usepackage{subcaption}       

\usepackage{booktabs}         
\usepackage{tabularx}         
\usepackage{multirow}         
\usepackage{array}            
\usepackage{adjustbox}        
\usepackage{siunitx}          

\usepackage{algorithm}        
\usepackage{algpseudocode}    
 
\usepackage{import}
\usepackage{xcolor}           
\usepackage{enumitem}         
\usepackage{lipsum}           
\usepackage{todonotes}        
\usepackage{appendix}         

\DeclareMathOperator*{\argmax}{arg\,max}

\usepackage{comment}

\usepgfplotslibrary{groupplots}
\usepackage{pgfplotstable}

\makeatletter
\newenvironment{customlegend}[1][]{%
    \begingroup
    \pgfplots@init@cleared@structures
    \pgfplotsset{#1}%
}{%
    \pgfplots@createlegend
    \endgroup
}%
\def\addlegendimage{\pgfplots@addlegendimage}

\definecolor{attrcol}{RGB}{31,119,180}    
\definecolor{objcol}{RGB}{255,127,14}     
\definecolor{paircol}{RGB}{44,160,44}     

\tikzset{
  attrcurve/.style={draw=attrcol, very thick, smooth},
  objcurve/.style ={draw=objcol,  very thick, smooth},
  paircurve/.style={draw=paircol, very thick, smooth},
  meanattr/.style ={attrcol, dashed, thick},
  meanobj/.style  ={objcol,  dashed, thick},
  meanpair/.style ={paircol, dashed, thick},
}
\pgfplotsset{
  attrcurve/.style={draw=attrcol, very thick, smooth},
  objcurve/.style ={draw=objcol,  very thick, smooth},
  paircurve/.style={draw=paircol, very thick, smooth},
  meanattr/.style ={attrcol, dashed, thick},
  meanobj/.style  ={objcol,  dashed, thick},
  meanpair/.style ={paircol, dashed, thick},
}

\definecolor{darkviolet}{rgb}{0.58, 0.0, 0.83}

\definecolor{wacvblue}{rgb}{0.21,0.49,0.74}
\usepackage[pagebackref,breaklinks,colorlinks,allcolors=wacvblue]{hyperref}

\def\wacvPaperID{91} 
\def\confName{WACV}
\def\confYear{2027}

\title{Compositionality with Primitives: \\ A Unified View from Euclidean to Hyperbolic Vision-Language Models}

\author{
    Puru Vaish$^{1}$ \quad
    Meelad Dashti$^{1,3}$ \quad
    Christoph Brune$^{1}$ \quad
    Pascal Mettes$^{2}$ \\
    Tatiana Tommasi$^{3}$ \quad
    Nicola Strisciuglio$^{1}$ \\[2mm]
    $^{1}$University of Twente, The Netherlands \quad
    $^{2}$University of Amsterdam, The Netherlands \\
    $^{3}$Politecnico di Torino, Italy 
}
\begin{document}

\maketitle

\begin{abstract}
Compositional zero-shot learning (CZSL) aims to recognize novel combinations of previously observed objects and attributes. Prior work showed that large-scale vision–language models encode decomposable semantic structures, that can be recomposed to enable generalization. 
The Geodesically Decomposable Embeddings (GDE) framework formalizes this by identifying  directions in the tangent space at the Fr\'{e}chet mean of the embedding distribution, which encode primitive visual concepts. Unseen attribute–object compositions can be synthesized through vector addition of these primitives.
We revisit the assumption that compositional generalization requires explicit attribute-object pair re-composition. By means of new compositional scoring functions and extensive evaluations, we show that directly using the decomposed primitives suffices for recognizing unseen compositions. In that, we gain insights on how the geometry of the embedding manifold (Euclidean, spherical, or hyperbolic) and dataset structure affect compositional reasoning.
Across models (CLIP, DINOv2, MERU, HyCoCLIP), we show that compositional structure consistently emerges as primitive representations,  enabling inference on CZSL benchmarks (MIT-States, VAW-CZSL, UT-Zappos50K, Clothing16K) and group robustness on Waterbirds and CelebA.
\footnote{Code: \href{https://github.com/nis-research/Compositionality-with-Primitives}{github.com/nis-research/Compositionality-with-Primitives}}
\end{abstract}

\section{Introduction}\label{sec:intro}

Recognizing a \emph{sliced banana} after seeing only \emph{sliced bread} and
\emph{whole bananas} requires disentangling the attributes \emph{sliced} and \textit{whole} from the
objects \textit{bread} and \emph{banana}, and reasoning over them independently. 
This ability reflects the broader principle of compositionality, a fundamental cognitive skill of humans clearly arising in language~\cite{lakeHumanlikeSystematicGeneralization2023}, whereby complex concepts are constructed from simpler, reusable ones~\cite{partee1995lexical}. At its core, it underpins the ability to generalize robustly to novel compositions without requiring direct supervision~\cite{lakeHumanlikeSystematicGeneralization2023}.

In Machine Vision, this paradigm is formalized as compositional zero-shot learning (CZSL)~\cite{Misra2017,munir2025compositionalzeroshotlearningsurvey}: given training images depicting a closed set of attribute-object pairs, the aim is to enable a model to recognize novel compositions of known concepts without having seen them during training.  Open-world CZSL~\cite{manciniOpenWorldCompositional2021} extended this setting to the full Cartesian product of attributes and objects, significantly increasing task
difficulty, and accounting for the long-tailed distribution of real-world generalization.
Existing solutions typically attempt to learn compositional representations using conditional formulations of attribute-object representations ~\cite{wu2025conditional,hu2023leveraging,Wang2023learningconditionalattributes}, or learning compositional soft-prompt~\cite{nayak2023learning,LuPrompt2023}, visual proxies~\cite{Zhang_2025_ICCV} or invariant representations~\cite{Zhang2022InvariantReprCompositional}.

%
%

Recent work on Geodesically Decomposable Embeddings
(GDE)~\cite{berasiNotOnlyText2025a} provides a different perspective: vision embeddings of pre-trained vision-language models already exhibit compositional semantic structures. These can be disentangled to perform compositional zero-shot classification. By projecting representations into the tangent space at the Fr\'echet mean of the embedding distribution of a dataset and performing a factorization step,
attribute and object representations emerge as vector directions that embed basic visual concepts, called primitives.  These can be recombined by vector arithmetics in the tangent space, thus synthesizing the representation of unseen object-attribute pairs and enabling their recognition.
However, the focus in~\cite{berasiNotOnlyText2025a} 
was limited to 
CLIP spherical embeddings~\cite{radfordLearningTransferableVisual2021}. 
It remains unclear whether this view of compositionality extends to other space geometries, i.e. spaces with flat (Euclidean) or negative (hyperbolic) curvature, and whether explicit composition of primitives into pairs is necessary to achieve open-world compositional generalization. This question gains particular relevance for hyperbolic models~\cite{desaiHyperbolicImagetextRepresentations2023,palCompositionalEntailmentLearning2025}, where embeddings lie on the Lorentz space and representations encode compositional hierarchy in a tree-like fashion.
%
%
\begin{figure}[t]
    \centering
    \input{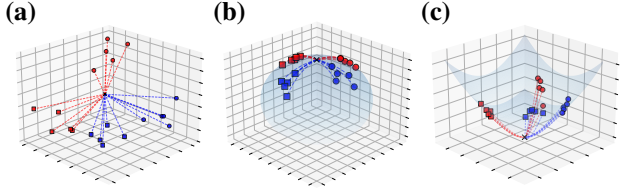}
    \vspace{-5ex}
    \caption{Geodesics initiating from the origin across Euclidean, spherical, and hyperbolic manifolds. The plots illustrate the structural differences imposed by space curvature: (a) \textbf{Euclidean space} is flat; (b) \textbf{spherical space} has positive curvature with origin estimated on the Fr\'echet mean of the dataset embeddings;  (c) \textbf{Hyperbolic space} has negative curvature with a well-defined origin, resulting in tree-like hierarchical representation space. Data clusters are differentiated by color and marker type (\{\textit{red}, \textit{blue}\} $\times$ \{\textit{square}, \textit{circle}\}).}
    \label{fig:geometries}
    \vspace{-3ex}
\end{figure}

In this work, we place primitives at the center of the analysis and revisit compositional reasoning in VLMs. By means of novel compositional scoring functions, we demonstrate that primitives can be used directly to perform compositional inference without the need of explicitly constructing (unseen) pair representations. We demonstrate improved open-world compositional generalization, 
across Euclidean, spherical, and hyperbolic representation
spaces, which hold different geometric and data distribution properties (see Figure~\ref{fig:geometries}). 
To perform this extensive investigation, we build upon the GDE framework and leverage its formulation beyond the original spherical setting of CLIP~\cite{radfordLearningTransferableVisual2021},
to a broad family of models across geometries. We investigate Euclidean DINOv2~\cite{barsellottiTalkingDINOBridging2024,maximeoquabDINOv2LearningRobust2023} 
and hyperbolic VLMs (MERU~\cite{desaiHyperbolicImagetextRepresentations2023}, HyCoCLIP~\cite{palCompositionalEntailmentLearning2025}), by utilizing corresponding manifold mapping functions between these geometries and the tangent space. 

We also develop a formulation of primitives directly on the embedding manifold as the Fr\'echet means of samples sharing an attribute or object, and call them \textit{intrinsic primitives}. We thus study manifold-native primitives as well as tangent-space GDE primitives, investigating how latent space geometry influences compositional abilities of models. 
%

Across experiments, we find that compositional structure is a general emergent property of pre-trained models, showing through both GDE and intrinsic primitive representations across several CZSL benchmarks (MIT-States, VAW-CZSL,
Clothing16K, UT-Zappos50K), with stronger evidence for datasets that contain samples with higher semantic variance. 
We further show that primitive-based inference is beneficial for improving recognition under group bias in the data through experiments on benchmark datasets for group robustness (Waterbirds, CelebA) using solely the object primitives, decoupled from the attribute primitives. 
The observed substantial and consistent increase of worst group accuracy indicates that
biases and spurious correlations induced by attributes can be removed with primitive-based reasoning.
%
%
%
%
%
%
%
%
%
%
The contributions of this work are therefore summarized as:
\begin{enumerate}[label=(\roman*),leftmargin=4ex,itemsep=0em]
%
    \item \textbf{Primitives as the basis for open-world compositional learning and generalization.}
    We show that pretrained models consistently organize representation into primitives, either as tangent-space directions or intrinsic points on manifolds, which encode basic visual concepts of attributes and objects. We develop new compositional scoring functions and demonstrate with extensive experiments that primitives support direct CZSL inference and open-world generalization, making explicit attribute-object pair construction unnecessary.
%
    \item \textbf{Understanding of compositional power of pretrained models in relation to manifold geometry and dataset intrinsic variance. }
    We find that, while compositional structure is geometry-agnostic, its task-wise optimal representation depends on the underlying geometry: tangent-space primitives are sufficient in Euclidean and spherical spaces, whereas hyperbolic representations benefit from intrinsic primitives aligned with their space geometry. We further shed light on the role that CZSL datasets and their intrinsic variance have in factorizing compositional primitives in pretrained models.
\end{enumerate}



\section{Related Work}\label{sec:related}

\noindent\textbf{Compositional zero-shot learning in VLMs.}
Early compositional zero-shot learning (CZSL) methods modeled attribute-object relations through shared embedding
spaces~\cite{isolaDiscoveringStatesTransformations2015,yuFinegrainedVisualComparisons2014}.
More recent studies have shifted attention to 
the compositional properties of Vision-Language models and highlighted that contrastive VLMs inherently struggle with text-based composition during standard inference, often behaving as a ``bag of words''~\cite{thrushWinogroundProbingVision2022,yuksekgonulWhenWhyVisionLanguage2023}.
This vulnerability is further underscored by benchmarks like SugarCrepe~\cite{hsiehSugarCrepeFixingHackable2023}, which demonstrate that VLMs frequently rely on statistical hacks rather than genuine compositionality.
In order to improve the compositional reasoning abilities of VLMs, 
subsequent research has explored  adaptation strategies based on soft-prompt tuning~\cite{nayak2023learning,LuPrompt2023}, or learning conditional attribute-object representations, where attribute frequency is coupled with the most likely co-occurring objects~\cite{wu2025conditional,hu2023leveraging,Wang2023learningconditionalattributes}. 

In~\cite{berasiNotOnlyText2025a}, it was hypothesised that the pre-trained vision encoder of VLMs natively encodes compositional structures, induced by language-based pre-training. Thus,  Geodesically Decomposable Embeddings (GDE) was proposed to factorize primitive representation directions in the tangent space at the Fr\'{e}chet mean of the 
embeddings distribution, 
enabling training-free compositional reasoning via constructing unseen combination of primitives. The focus was 
restricted to spherical (CLIP-like) embeddings and to explicit construction of object-attribute pair representations for CZSL. 

In this work, we fundamentally revisit the assumption that explicit pair re-composition is necessary for CZSL inference. We formulate new scoring functions for compositional classification that operate directly on independent object and attribute primitives. We show that leveraging these primitive representations is sufficient to 
recognize unseen compositions, bypassing pair recombination entirely. Furthermore, we expand the GDE framework to work across different geometries of the latent space (i.e. Euclidean, spherical and hyperbolic) and demonstrate the emergence of compositional structures in a broad family of visual encoders.

Our work differs from fine-tuning~\cite{nayak2023learning,LuPrompt2023} and conditional attribute learning~\cite{wu2025conditional,hu2023leveraging,Wang2023learningconditionalattributes}, as we do not add learnable components and study the embedding capabilities of pretrained VLMs.
%
It also shares a philosophical connection with Concept Bottleneck Models (CBMs)~\cite{kohConceptBottleneckModels2020}, which enforce intermediate, primitive-based concept representations for downstream inference. However, while CBMs require explicit bottleneck layers or fine-tuning with concept supervision, we analyze the compositional structure that inherently emerges within the embedding space of pretrained models. 

%
%


\smallskip
\noindent\textbf{Compositionality in hyperbolic VLMs.}
Most vision-language models operate on the unit hypersphere using cosine
similarity.
In contrast, hyperbolic models
embed representations on the Lorentz model~\cite{palCompositionalEntailmentLearning2025,desaiHyperbolicImagetextRepresentations2023,ramasingheAcceptModalityGap2024,poppiHyperbolicSafetyawareVisionlanguage2025,yoshikawaPHyCLIPproductHyperbolic2026}, where radius reflects semantic specificity, and geometry captures hierarchical structure: abstract concepts are placed closer to the origin, while more complex detailed concepts lie further away. 
%
HyCoCLIP~\cite{palCompositionalEntailmentLearning2025} further incorporates
compositional entailment supervision to align scene-object 
compositions with the hierarchical organization of the embedding space. 
While prior work has primarily focused on retrieval, entailment, and hierarchy preservation, we instead investigate how hyperbolic geometry shapes compositional reasoning.
We show that primitives naturally align with the manifold structure, and that for group robustness intrinsic primitives provide a more semantically-coherent representation than tangent-space approximations in hyperbolic settings.

\smallskip
\noindent\textbf{Compositionality in vision-only DINO.}
Compositional capabilities are not exclusive to cross-modal vision-language models. Self-supervised vision-only foundation models, such as DINOv2~\cite{maximeoquabDINOv2LearningRobust2023}, capture exceptional fine-grained, localized spatial semantics without using any language-signal during training. Recent efforts 
have explored post-hoc mechanisms to 
interface language with these frozen visual encoders. For instance, Talk2DINO~\cite{barsellottiTalkingDINOBridging2024} demonstrated that text embeddings can be aligned, via a learned projection, 
onto DINO patch-level features to guide open-vocabulary tasks. 
These findings suggest that a structured semantic organization emerges within the token embeddings of vision-only pre-trained encoders. 
Understanding the geometry underlying this intrinsic compositionality is therefore crucial, as such models provide powerful representations that are not influenced by language biases.
In this work, we apply the GDE framework to DINOv2 and show that compositional structure naturally emerges in its (Euclidean) latent space, despite the absence of language supervision.

\setlength{\belowdisplayskip}{6pt} \setlength{\belowdisplayshortskip}{6pt}
\setlength{\abovedisplayskip}{6pt} \setlength{\abovedisplayshortskip}{6pt}

\section{Preliminaries}\label{sec:prelim}
In this section, we provide a brief overview of the mathematical background relevant to the notion of compositional embeddings and the GDE framework. Moreover, we provide the formalization of open-world CZSL and pair-based scoring functions used for recognition, and cover the foundational alignment of DINOv2 patch embeddings with text embeddings of CLIP done via the Talk2DINO framework.

%

\begin{figure*}[t]
    \centering
    \vspace{-3ex}
    \def\svgwidth{0.70\linewidth}
    \import{figures/schema}{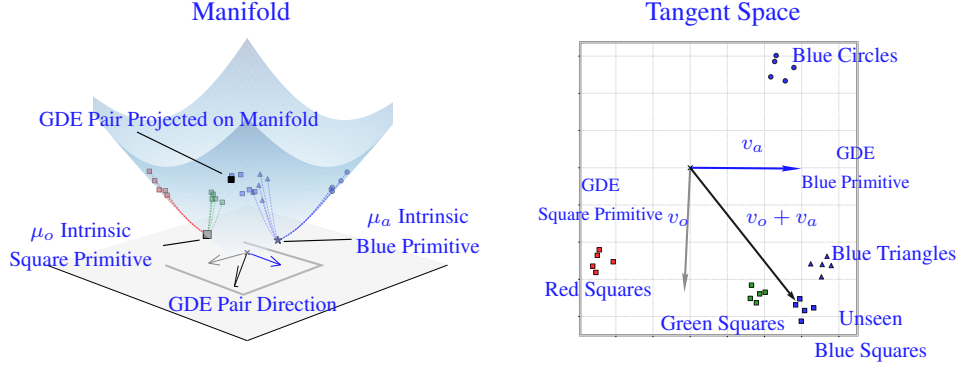}
    \caption{
        Illustration of decomposition and primitive-based methods. 
        For unseen samples of \emph{Blue Squares}, the goal is to obtain meaningful representations of the concepts \emph{Blue} and \emph{Square} from seen training samples. 
        Then, the primitive directions in the tangent space ($\textbf{v}_o$ and $\textbf{v}_a$) can independently be compared to the test samples projected to the same space. 
        Alternatively, it is possible to calculate the intrinsic mean for each concept label directly on the manifold ($\mu_a$ and $\mu_o$ and use it as a reference for the test sample via a manifold-specific inner product to estimate similarity. 
        The pair-based alternatives proposed in previous work either rely on the projection on the manifold of the summed primitives in the tangent space ($\textbf{v}_o+\textbf{v}_a$), or leverage the textual object-attribute compositional description embedded in the manifold.
    }
    \label{fig:inference_styles}
    \vspace{-1ex}
\end{figure*}

\subsection{Geodesically Decomposable Embeddings}
VLMs encode images $x \in \mathcal{X}$ and their associated caption $y\in\mathcal{Y}$ into shared embedding spaces via specific encoders $\phi_{im}:\mathcal{X} \rightarrow \mathcal{R}^d$, $\phi_{tx}:\mathcal{Y} \rightarrow \mathcal{R}^d$. Let $\mathcal{M}\subset\mathcal{R}^d$ be the target $d$-dimensional representation manifold, with $\mathbf{u}\in\mathcal{M}$ being a generic point on the manifold. The text prompts can be created by a general template function $g: \mathrm{L}^n \to \mathcal{Y} \quad \forall n \in \mathbb{N}$ where $\mathrm{L}^n$ are the $n$ concepts to make a template for.

%
In the Geodesically Decomposable Embeddings (GDE) framework~\cite{berasiNotOnlyText2025a}, visual representations from pre-trained VLMs are decomposed into primitive directions corresponding to objects and attributes, which can then be recombined for CZSL.
Given $N$ samples belonging to a dataset, their intrinsic (Fr\'echet) mean is 
${\boldsymbol{\mu} =
    \arg\min_{\mathbf{u}\in\mathcal{M}}
    \sum_{i=1,\ldots N}
    d_{\mathcal{M}}(\mathbf{u},\mathbf{u}_i)^2}$,
where $d_{\mathcal{M}}$ denotes the geodesic distance on the manifold. Generally, a Riemannian gradient descent approach with a finite number of iterations is employed to compute the Fr\'echet mean on any manifold geometry (see Algorithm~\ref{alg:frechet_mean} in the supplementary material). Let $T_{\boldsymbol{\mu}}\mathcal{M}$ be the tangent space to the embedding manifold at $\boldsymbol{\mu}$, and $\mathbf{t}_i = \mathrm{Log}_{\boldsymbol{\mu}}(\mathbf{u}_i) \in T_{\boldsymbol{\mu}}\mathcal{M}$ the projection of an embedding vector to the tangent space via the \emph{logarithmic map}. The function is locally invertible, and its inverse is the \emph{exponential map}  $\mathbf{u}_i = \mathrm{Exp}_{\boldsymbol{\mu}}(\mathbf{t}_i) \in \mathcal{M}$.  
The exact form of the map functions depends on the manifold geometry detailed in Section~\ref{sec:method}.

Given a set of attributes $\mathcal{A} = \{a_1, a_2, \dots, a_n \}$ and objects
$\mathcal{O} = \{o_1, o_2, \dots, o_m \}$, GDE estimates an independent,
isolated primitive semantic direction $\mathbf{v}_c$ for any concept
$c \in \mathcal{A} \cup \mathcal{O}$ via a weighted average of the projected
tangent vectors containing that concept:
\begin{align}
    \mathbf{v}_c
    =
    \frac{
        \sum_{i \in S_c}
        w_i \mathbf{t}_i
    }{
        \sum_{i \in S_c}
        w_i
    },
    \qquad
    S_c
    =
    \{ i : y_i = c \},
    \label{eq:gde_weighted}
\end{align}
where $y_i$ denotes the attribute or object label associated with sample $i$,
and each sample is labelled by a compositional pair
$(a_i,o_i)\in\mathcal{A}\times\mathcal{O}$.
Following~\cite{berasiNotOnlyText2025a}, each embedding is weighted according
to its alignment with the corresponding attribute--object text representation
\[
    w
    \propto
    \exp\!\Big(
        \tau\,
        s_{\mathcal{M}}\big(
            \phi_{im}(x),
            \phi_{tx}(g(a,o))
        \big)
    \Big),
\]
where $\tau$ is a temperature parameter and
$s_{\mathcal{M}}(\cdot,\cdot)$ is a similarity function defined as a
geometry-specific inner product.

\subsection{Open-World CZSL and Pair-based Scoring}
\label{sec:CZSL_and_pair}
%
In the open-world CZSL (OW-CZSL) setting~\cite{manciniOpenWorldCompositional2021}, 
models are evaluated over the full Cartesian product ${\mathcal{A}\times\mathcal{O}}$ of possible compositions at test
time, regardless of which attribute-object pairs were observed during training.
This removes the disambiguation advantage of closed candidate sets and
provides a stronger test for compositional generalization.
For the zero-shot inference baseline, we prompt VLMs using a pre-defined text template in the form of ~`\verb!a photo of an {attribute} {object}!', which is embedded via the text encoder and used to score the vision embeddings 
$\phi_{im}(x)$.
%
%
The scores computed using the function $s_{\mathcal{M}}$  are conventionally converted into probabilities through a softmax over the valid label space. 
The baseline \textbf{Zero-Shot Pair} paradigm is based on the scoring function (Eq.~\ref{eq:zs-pair}),
%
which calculates 
the probability assigned to an attribute-object pair and the prediction is obtained by selecting the label with max-probability (Eq.~\ref{eq:prob}). 
\begin{align}
    s(a, o | x) &= s_{\mathcal{M}}(\phi_{im}(x), \phi_{tx}(g(a,o))),
    \label{eq:zs-pair} \\
\begin{split}
    (\hat{a},\hat{o})
    &=\argmax_{(a,o)\in\mathcal{A}\times\mathcal{O}}
    p(a,o \mid x),
    \\
    p(a,o \mid x)
    &=
    \frac{
        \exp(s(a, o | x))
    }{
        \sum_{a' \in \mathcal{A}}
        \sum_{o' \in \mathcal{O}}
        \exp(s(a', o' | x))
    }.
    \label{eq:prob}
\end{split}
\end{align}
Rather than using the text prompts, the GDE paper \cite{berasiNotOnlyText2025a} proposed to combine all possible attribute-object primitive pairs and re-project them from the tangent space to the embedding manifold. For the \textbf{GDE Pair} paradigm, compositional recognition is achieved by using the scoring function: 
\begin{equation}
s(a, o | x) = s_{\mathcal{M}}(\phi_{im}(x), \mathrm{Exp}_{\boldsymbol{\mu}}(\mathbf{v}_a+\mathbf{v}_o))
\label{eq:gde-pair}.
\end{equation}
%
Here we raise the question of whether exhaustive pair recombination of primitives is necessary for OW-CZSL or if scores derived from single concept primitive are sufficient and perhaps better instead. However, we must first define scoring with primitives and their computation.

\section{Primitive-based CZSL}
\label{sec:method}
The core focus of this work is the representation of visual concept as primitives (i.e. isolated attributes and objects representations), that form fundamental building blocks of compositional generalization. Rather than evaluating compositionality via pre-defined textual or visual pairs, we investigate how individual primitives emerge within different manifold geometries. 
Specifically, we study two distinct approaches for extracting these representations: \emph{intrinsic primitives} computed directly on the underlying manifold via Riemannian operations, and \emph{tangent primitives} extracted as localized linear vector directions via the Geodesically Decomposable Embeddings (GDE) framework. Under both paradigms, inference is performed zero-shot by scoring test images directly against these standalone primitives in their native mathematical domains. This eliminates the need for explicit primitive re-composition, currently used as standard in GDE. Figure~\ref{fig:inference_styles} illustrates the differences between intrisic primitives, GDE tangent primitives, and GDE-Pair compositions on the hyperbolic geometry as a representative case.

\subsection{Intrinsic Primitives via Fr\'echet Means}\label{sec:int_prim}
We define an \emph{intrinsic primitive}
as the manifold mean of samples sharing the same attribute or object label.
Given a set of embeddings $\{f(x_i)\}_{i=1}^N \subset \mathcal{M}$, and a concept $c$, it is computed as the Fr\'echet mean of the
subset, $S_c$, associated with that concept:
\begin{equation}
    \boldsymbol{\mu}_{c}
    =
    \arg\min_{\mathbf{u}\in\mathcal{M}}
    \sum_{i \in S_{c}} d_{\mathcal{M}}^2(\mathbf{u}, \phi_{im}(x_i)),
\end{equation}
where $d_{\mathcal{M}}$ denotes the
geodesic distance on the manifold.
We compute separate intrinsic primitives for attributes and objects, using the Riemannian gradient descent for non-Euclidean spaces and the arithmetic mean for Euclidean ones, 
yielding respectively 
$\boldsymbol{\mu}_a$ for each attribute $a \in \mathcal{A}$ and
$\boldsymbol{\mu}_o$ for each object $o \in \mathcal{O}$.
%
%

Given a test image $x$, denoting its embedding as $\phi^x_{im}$ for brevity, its similarity to the primitives is computed directly 
on the underlying
manifold using the geometry-specific similarity function
$s_{\mathcal{M}}$.
In particular, for a reference pair of object ($o$) and attribute ($a$) and text template $g_c := g(c) \quad c\in \{a, o\}$ with its corresponding text embedding $\phi^c_{tx}$, the \textit{Zero Shot Primitive} score is defined as:
\begin{align}
s^{\mathrm{zsp}}(a,o \mid \cdot) = s_{\mathcal{M}}(\phi^x_{im}, \phi^a_{tx})
 + s_{\mathcal{M}}(\phi^x_{im}, \phi^o_{tx}).
\label{eq:zs-prim}
\end{align}
In contrast, our scoring function that uses \textit{intrinsic primitive} individually does not rely on the text prompt embeddings:
\begin{align}
    s^{\mathrm{intrp}}(a,o \mid \cdot) &= s_{\mathcal{M}}(\phi^x_{im}, \boldsymbol{\mu}_a)
 + s_{\mathcal{M}}(\phi^x_{im}, \boldsymbol{\mu}_o).
\label{eq:intr-prim}
\end{align}
%

\noindent The similarity functions used for the underlying spherical, Euclidean and hyperbolic manifolds are respectively 
\begin{align*}
s_{\mathbb{S}^{d-1}}(\mathbf{x},\mathbf{y})
=
\mathbf{x}^\top\mathbf{y},
\;\;\;\;&\;\;\;\;
s_{\mathbb{R}^d}(\mathbf{x},\mathbf{y})
=
\frac{
    \mathbf{x}^\top\mathbf{y}
}{
    \|\mathbf{x}\|\,\|\mathbf{y}\|
},
\\
s_{\mathbb{H}^{d+1}}(\mathbf{x},\mathbf{y})
&=
\sum_{k=1}^{d}x_ky_k
-
x_0y_0.
\end{align*}

\subsection{GDE Primitives on any Manifold}
We exploit the GDE framework on different manifold geometries by integrating the logarithmic map $\mathrm{Log}_{\boldsymbol{\mu}}: \mathcal{M} \to T_{\boldsymbol{\mu}}\mathcal{M}$ and the exponential map $\mathrm{Exp}_{\boldsymbol{\mu}}: T_{\boldsymbol{\mu}}\mathcal{M} \to \mathcal{M}$ specifically defined for spherical, Euclidean and Lorentz spaces.

For \textbf{Euclidean manifolds} $\mathbb{R}^d$, the tangent space at every point is naturally identified with the manifold itself. In this case, the logarithmic and exponential maps reduce to ordinary subtraction and addition:
\begin{align}
    \mathrm{Log}_{\boldsymbol{\mu}}(\mathbf{u}) = \mathbf{u}-\boldsymbol{\mu} && \mathrm{Exp}_{\boldsymbol{\mu}}(\mathbf{t})= \boldsymbol{\mu}+\mathbf{t}.
\end{align}

For \textbf{spherical manifolds} $\mathbb{S}^{d-1}$ , the logarithmic and exponential maps are:
\begin{align}
\begin{split}
  \mathrm{Log}_{\boldsymbol{\mu}}(\mathbf{u})
    &=
    \arccos(\boldsymbol{\mu}^\top\mathbf{u})
    \frac{
      \mathbf{u} - (\boldsymbol{\mu}^\top\mathbf{u})\boldsymbol{\mu}
    }{
      \|\mathbf{u} - (\boldsymbol{\mu}^\top\mathbf{u})\boldsymbol{\mu}\|
    },
  \\
  \mathrm{Exp}_{\boldsymbol{\mu}}(\mathbf{t})
    &=
    \cos(\|\mathbf{t}\|)\boldsymbol{\mu}
    +
    \sin(\|\mathbf{t}\|)
    \frac{\mathbf{t}}{\|\mathbf{t}\|}~.
  \label{eq:sphere_maps}
\end{split}
\end{align}

For \textbf{hyperbolic models}, with embeddings lying on the Lorentz space, $\mathbb{H}^{d+1}$ the logarithmic and exponential maps are:
\begin{align}
\begin{split}
  \mathrm{Log}_{\boldsymbol{\mu}}(\mathbf{u})
    &=
    \frac{
      \mathrm{arccosh}\big(
      -\langle\boldsymbol{\mu},\mathbf{u}\rangle_{\mathcal{L}}
      \big)
    }{
      \sqrt{
      \langle\boldsymbol{\mu},\mathbf{u}\rangle_{\mathcal{L}}^2 - 1
      }
    }
    \Big(
      \mathbf{u}
      +
      \langle\boldsymbol{\mu},\mathbf{u}\rangle_{\mathcal{L}}\boldsymbol{\mu}
    \Big),
  \\
  \mathrm{Exp}_{\boldsymbol{\mu}}(\mathbf{t})
    &=
    \cosh(\|\mathbf{t}\|_{\mathcal{L}})\boldsymbol{\mu}
    +
    \sinh(\|\mathbf{t}\|_{\mathcal{L}})
    \frac{\mathbf{t}}{\|\mathbf{t}\|_{\mathcal{L}}}, \\ 
    \text{where} \quad \|\mathbf{t}\|_{\mathcal{L}}
    &=
    \sqrt{\langle\mathbf{t},\mathbf{t}\rangle_{\mathcal{L}}}~.
\label{eq:hyp_maps}
\end{split}
\end{align} 

\noindent The similarity functions for each geometry are the same defined at the end of Section~\ref{sec:int_prim}.

The primitive directions $(\mathbf{v}_a, \mathbf{v}_o)$, obtained via the GDE framework, and the projected test image representation 
$\mathbf{t}_x = \mathrm{Log}_{\boldsymbol{\mu}}(\phi_{im}(x))$ 
all lie in the same tangent space $T_{\boldsymbol{\mu}}\mathcal{M}$, where relationships are linear.
We thus, in contrast to Eq.~\ref{eq:gde-pair}, define scoring functions for CZSL using individual \textit{GDE primitives}, in which attribute and object scores are computed via standard cosine similarity in the tangent space as:
\begin{align}
    s^{\mathrm{tan}}(a,o \mid x)
    &=
    s_{\mathbb{R}^d}(\mathbf{t}_x,\mathbf{v}_a)
    +
    s_{\mathbb{R}^d}(\mathbf{t}_x,\mathbf{v}_o),
    \label{eq:gde-prim}
\end{align}
bypassing manifold projections entirely during inference.

%

\section{Experiments}\label{sec:experiment}

\subsection{Setup}

\noindent\textbf{Datasets.}
We evaluate compositional recognition on four datasets spanning varying levels of semantic diversity and attribute-object coupling.
MIT-States~\cite{isolaDiscoveringStatesTransformations2015} contains 115 attributes paired with 245 objects, yielding 28,175 open-world compositions.
VAW-CZSL~\cite{SainiVAW22,phamLearningPredictVisual2021} is substantially larger, with 440 attributes and 541 objects (238,140 candidate compositions) sampled from images with larger scene context.
UT-Zappos~\cite{yuFinegrainedVisualComparisons2014} is focused on footwear with 16 material attributes and 12 shoe types, while Clothing16K provides a broader clothing-domain benchmark with greater intra-category variability.
Together, these datasets allow us to study compositionality under both highly diverse semantic settings (MIT-States, VAW-CZSL) and more constrained cases where sample variance is small and attributes are tightly coupled to object identity (UT-Zappos, Clothing16K).

\noindent\textbf{Models and baselines.}
We investigate models operating in Euclidean, spherical, and hyperbolic embedding spaces.
Spherical models include CLIP: ViT-B/16 (GRIT dataset) and ViT-L/14 (OpenAI).
For Euclidean representations, we use DINOv2 with Talk2DINO~\cite{barsellottiTalkingDINOBridging2024} bridge to the text embedding of CLIP. This preserve the native Euclidean embeddings of DINOv2 and enables to compute baseline results with CLIP-style text prompts (see Appendix~\ref{app:dino_dense}).
We also experiment with MERU~\cite{desaiHyperbolicImagetextRepresentations2023} and
HyCoCLIP~\cite{palCompositionalEntailmentLearning2025} hyperbolic models.
CLIP-B, MERU-B, and HyCoCLIP-B share the same ViT backbone and are pretrained on the GRIT corpus~\cite{pengKosmos2GroundingMultimodal2023}, thus allowing to attribute performance differences primarily to the underlying manifold geometry.

\noindent\textbf{Prompts and inference paradigms.}
We test the compositional zero-shot learning performance of the considered vision-language models under different settings, namely:
\begin{enumerate}[label=(\roman*),leftmargin=4ex,itemsep=0em]
    \item \textbf{ZS-Pair}: direct zero-shot classification using pair-based text prompts ~\texttt{A photo of an \{attribute\} \{object\}} as defined in Eq.~\ref{eq:zs-pair}; 
    \item \textbf{ZS-Prim}: primitive-level zero-shot inference using text prompts with attribute-focus (\texttt{A photo of an \{attribute\}}) and object-focus (\texttt{A photo of an \{object\}}) using the scoring function defined in Eq.~\ref{eq:zs-prim};
    \item \textbf{GDE-Pair}: composing pair representations by adding visual primitives in the tangent space as $\mathbf{v}_{a,o}=\mathbf{v}_a+\mathbf{v}_o$, and scoring the composed vector as defined in~\cite{berasiNotOnlyText2025a};
    \item \textbf{GDE-Prim}: scoring the GDE object and attribute primitives independently using our function defined in Eq.~\ref{eq:gde-prim};
    \item \textbf{Int-Prim}: scoring the intrinsic object and attribute primitives independently  using our function defined in Eq.~\ref{eq:intr-prim}.
    
\end{enumerate}

\noindent\textbf{Metrics.} 
We compute the \textit{unseen pair accuracy} (UPA) that evaluates the generalization to novel attribute-object compositions, and the \textit{seen pair accuracy} (SPA) that measures the recognition performance on compositions observed during training (we report SPA in Appendix~\ref{app:extended_metrics}). 
Models often exhibit a bias toward seen compositions, especially in the open-world setting where all feasible compositions are considered candidate predictions. Evaluating only one regime can provide a misleading picture of overall performance. Hence, as standard practice in OW-CZSL~\cite{manciniOpenWorldCompositional2021}, we report the best harmonic mean of \emph{seen} and \emph{unseen} compositions:
\[
\mathrm{HM} = \frac{2 \cdot \mathrm{SPA} \cdot \mathrm{UPA}}{\mathrm{SPA} + \mathrm{UPA}},
\]
that penalizes models that perform well on only one subset and therefore captures the ability to maintain strong recognition on both seen and unseen compositions simultaneously.

\subsection{Results: CZSL via Primitives}
\label{sec:main_results}

\noindent\textbf{Primitives improve open-world generalization.}
\begin{table*}[t]
\centering
\caption{Harmonic Mean (HM) and Unseen Pair Accuracy (UPA) results across models and datasets. Primitive-based inference consistently improves compositional generalization over GDE pair composition, particularly on unseen attribute--object pairs.}
\vspace{-1mm}
\label{tab:hm_pair}
\resizebox{\linewidth}{!}{%
    \begin{tabular}{
@{}ll
ccccc
ccccc@{}
}
\toprule

& &
\multicolumn{5}{c}{\textbf{Harmonic Mean (\%, $\uparrow$)}} &
\multicolumn{5}{c}{\textbf{Unseen Pair Accuracy (\%, $\uparrow$)}} \\
\cmidrule(lr){3-7}
\cmidrule(lr){8-12}
 &
\textbf{Model} &
\textbf{ZS Pair} &
\textbf{ZS Prim.} &
\textbf{GDE Pair} &
\textbf{GDE Prim.} &
\textbf{Int. Prim.} &
\textbf{ZS Pair} &
\textbf{ZS Prim.} &
\textbf{GDE Pair} &
\textbf{GDE Prim.} &
\textbf{Int. Prim.} \\
\midrule
\multirow{5}{*}{\makebox[0.5cm]{\rotatebox[origin=c]{90}{MIT-States}}}
& CLIP ViT-L/14
    & 8.03 & 10.56 & 9.37 & \textbf{10.90} & 10.40
    &  8.04 & \textbf{13.03} &  8.62 & 11.98 & 11.75 \\
& DINOv2 ViT-L/14
    & 3.00 & 3.69 & 6.56 & \textbf{8.22} & 8.04
    &  2.88 & 4.78 &  6.26 &  9.33 &  \textbf{9.42} \\
& CLIP-B (GRIT)
    & 4.61 & 4.70 & 4.75 & \textbf{6.64} & 6.38
    &  4.24 &  5.36 &  4.80 &  \textbf{7.18} &  7.12 \\
& MERU-B
    & 4.68 & 4.74 & 3.58 & \textbf{6.57} & 6.36
    &  4.40 &  5.54 &  3.40 &  7.25 &  \textbf{7.40} \\
& HyCoCLIP-B
    & 5.26 & 6.53 & 4.28 & \textbf{7.94} & 7.54
    &  4.67 &  8.20 &  3.71 &  8.75 &  \textbf{8.76} \\
\midrule
\multirow{5}{*}{\makebox[0.5cm]{\rotatebox[origin=c]{90}{VAW-CZSL}}}
& CLIP ViT-L/14    & 0.74 & 1.51 & 1.18 & \textbf{1.77} & \textbf{1.77}
                   & 0.79 & 1.59 & 1.32 & 2.07 & \textbf{2.22} \\
& DINOv2 ViT-L/14  & 0.72 & 1.22 & 1.30 & 2.05 & \textbf{2.20}
                   & 0.77 & 1.57 & 1.63 & 2.60 & \textbf{2.81} \\
& CLIP-B (GRIT)    & 0.26 & 0.72 & 0.96 & \textbf{1.52} & 1.49
                   & 0.41 & 0.96 & 1.30 & \textbf{2.13} & 1.95 \\
& MERU-B           & 0.36 & 0.77 & 0.86 & \textbf{1.49} & 1.34
                   & 0.49 & 0.96 & 1.00 & \textbf{1.97} & 1.83 \\
& HyCoCLIP-B       & 0.29 & 1.14 & 0.85 & \textbf{1.47} & 1.40
                   & 0.33 & 1.20 & 1.08 & \textbf{2.09} & 1.87 \\
\midrule
\multirow{5}{*}{\makebox[0.5cm]{\rotatebox[origin=c]{90}{Clothing16k}}}
& CLIP ViT-L/14    & 91.67 & 85.33 & 58.81 & \textbf{92.23} & 92.11
                   & 94.84 & 91.26 & 63.77 & \textbf{95.40} & 95.20 \\
& DINOv2 ViT-L/14  & 21.89 & 27.09 & 13.95 & \textbf{33.71} & 33.37
                   & 22.34 & 34.98 & 15.95 & 43.51 & \textbf{44.94} \\
& CLIP-B (GRIT)    & 87.64 & 79.19 & 64.60 & \textbf{89.27} & 88.92
                   & 89.94 & 83.36 & 72.17 & \textbf{93.22} & 92.59 \\
& MERU-B           & 86.25 & 76.99 & 67.57 & \textbf{88.73} & 88.07
                   & 90.14 & 83.09 & 74.35 & 92.32 & \textbf{92.52} \\
& HyCoCLIP-B       & 87.01 & 87.29 & 70.47 & \textbf{88.84} & 88.31
                   & 89.18 & 91.13 & 75.31 & \textbf{92.89} & 92.52 \\
\bottomrule
\end{tabular}%
}
\vspace{-3mm}
\end{table*}
\begin{table}[t]
\centering
\caption{CZSL metrics before and after whitening across models and inference methods for the low-intrinsic variance  \textbf{UT-Zappos dataset}. Whitening consistently improves primitive-based inference and restores its advantage over pair composition.}
\label{tab:ut_zap}
\resizebox{\linewidth}{!}{%
\begin{tabular}{
@{}l
cccccc@{}}
\toprule
& \multicolumn{5}{c}{\textbf{Harmonic Mean (\%, $\uparrow$) / After Whitening}} \\ \midrule
\textbf{Model} &
\textbf{ZS Pair} &
\textbf{ZS Prim.} &
\textbf{GDE Pair} &
\textbf{GDE Prim.} &
\textbf{Int. Prim.} \\ \midrule
CLIP ViT-L/14       & 11.76 / -- & 9.61 / -- & 16.41 / 16.69 & 8.04 / 13.75 & 9.33 / \textbf{22.80} \\ 
DINOv2 ViT-L/14     & ~2.81 / -- & 0.38 / -- & 12.06 / 13.64 & 3.20 / 12.40 & 2.49 / \textbf{16.17} \\ 
CLIP-B (GRIT)       & ~2.33 / -- & 1.61 / -- & 10.81 / 13.53 & 9.04 / 12.79 & 8.95 / \textbf{14.65} \\ 
MERU-B              & ~3.38 / -- & 3.80 / -- & 10.39 / 12.79 & 8.99 / ~8.80 & 9.36 / \textbf{16.32} \\ 
HyCoCLIP-B          & ~4.34 / -- & 3.94 / -- & ~8.50 / 13.01 & 5.89 / ~8.14 & 5.86 / \textbf{15.11} \\ 
\midrule
& \multicolumn{5}{c}{\textbf{Unseen Pair Accuracy (\%, $\uparrow$) / After Whitening}} \\ \midrule
\textbf{Model} &
\textbf{ZS Pair} &
\textbf{ZS Prim.} &
\textbf{GDE Pair} &
\textbf{GDE Prim.} &
\textbf{Int. Prim.} \\ \midrule
CLIP ViT-L/14       & 21.89 / -- & 29.19 / -- & 29.46 / 36.81 & 18.03 / 24.43 & 22.85 / \textbf{36.83} \\ 
DINOv2 ViT-L/14     & ~9.84 / -- & 11.85 / -- & 21.95 / 28.34 & 12.64 / 19.78 & 16.08 / \textbf{28.42} \\ 
CLIP-B (GRIT)       & 16.45 / -- & 11.63 / -- & 16.50 / 21.01 & 13.11 / 15.44 & 13.91 / \textbf{21.21}  \\ 
MERU-B              & ~8.41 / -- & ~8.51 / -- & 16.45 / \textbf{25.17} & 11.26 / 15.55 & 14.60 / 25.12 \\ 
HyCoCLIP-B          & 11.95 / -- & 14.49 / -- & 14.65 / 23.69 & 12.16 / 16.08 & 14.01 / \textbf{24.80} \\ 
\bottomrule
\end{tabular}%

}
\vspace{-3mm}
\end{table}
In Table~\ref{tab:hm_pair} we reports the Harmonic Mean (HM) and Unseen Pair Accuracy (UPA) results of the CZSL performance of the considered pretrained models. 
Across models and datasets, replacing GDE pair composition with primitive-based inference consistently improves both metrics, indicating that explicit pair re-composition is not necessary for compositional generalization.
On MIT-States, GDE primitives improve HM over GDE pairs by +1.53 for CLIP ViT-L/14, +1.66 for DINOv2, +1.89 for CLIP-B, +2.99 for MERU-B, and +3.66 for HyCoCLIP-B.
The gains are often larger for UPA, including +3.85 for MERU-B and +5.04 for HyCoCLIP-B, indicating that primitives specifically strengthen open-world generalization.
Notably, DINOv2 also benefits from primitive decomposition despite weak zero-shot pair prediction performance, improving HM by +1.65 on MIT-States and +0.90 on VAW-CZSL.
As DINOv2 is trained without image-text contrastive supervision, this suggests that compositional semantics intrinsically emerge also from large-scale vision-only self-supervision.
Across geometries, intrinsic primitives provide the most stable formulation, particularly in hyperbolic space, consistently with representing visual concept directly on the embedding manifold rather than through tangent-space approximations.
Overall, primitive representations provide a stronger tool for open-world compositional generalization than explicit pair composition.

\begin{figure}[t]
    \centering
    \resizebox{\linewidth}{!}{
    \begin{tikzpicture}
\begin{groupplot}[
    group style={
        group size=3 by 1,
        horizontal sep=1.0cm,   
    },
    width=6cm, height=7cm,
    ymin=0,
    enlarge x limits=0.2,
    xtick=data,
    xticklabel style={rotate=45, anchor=east, font=\small},
    ylabel style={font=\small},
    xlabel style={font=\small},
    yticklabel style={
        font=\small,
        /pgf/number format/fixed,
        /pgf/number format/precision=2
    },
    scaled y ticks=false,
]
\nextgroupplot[
    title={DINOv2-L/14}, ylabel={Geometric Variance}, ybar, bar width=6pt, symbolic x coords={MIT-States,VAW-CZSL,Clothing-16k,UT-Zappos,UTZ-W}, xtick=data
]
\addplot[fill=blue!60!white] coordinates {(MIT-States,52.5859) (VAW-CZSL,54.9363) (Clothing-16k,38.2702) (UT-Zappos,22.5207) (UTZ-W,50.780)};
\addplot[fill=orange!80!white] coordinates {(MIT-States,29.4274) (VAW-CZSL,31.2829) (Clothing-16k,17.9137) (UT-Zappos,7.7893) (UTZ-W,32.175)};

\nextgroupplot[
    title={CLIP ViT-L/14}, ybar, bar width=6pt, symbolic x coords={MIT-States,VAW-CZSL,Clothing-16k,UT-Zappos,UTZ-W}, xtick=data
]
\addplot[fill=blue!60!white] coordinates {(MIT-States,0.2047) (VAW-CZSL,0.1517) (Clothing-16k,0.1275) (UT-Zappos,0.0681) (UTZ-W,0.143)};
\addplot[fill=orange!80!white] coordinates {(MIT-States,0.1298) (VAW-CZSL,0.1183) (Clothing-16k,0.0919) (UT-Zappos,0.0406) (UTZ-W,0.112)};

\nextgroupplot[
    title={HyCoCLIP-B}, ybar, bar width=6pt, symbolic x coords={MIT-States,VAW-CZSL,Clothing-16k,UT-Zappos,UTZ-W}, xtick=data
]
\addplot[fill=blue!60!white] coordinates {(MIT-States,0.5661) (VAW-CZSL,0.4165) (Clothing-16k,0.4066) (UT-Zappos,0.1304) (UTZ-W,0.498)};
\addplot[fill=orange!80!white] coordinates {(MIT-States,0.3866) (VAW-CZSL,0.3633) (Clothing-16k,0.2952) (UT-Zappos,0.0791) (UTZ-W,0.387)};

\end{groupplot}
\end{tikzpicture}%
    }

    \begin{tikzpicture}
    \begin{customlegend}[
        legend columns=-1,
        legend style={draw=none,column sep=1.5ex,font=\scriptsize},
        legend entries={
            Object Intrinsic Primitives Variance,
            Attribute Intrinsic Primitives
        }
    ]
    \addlegendimage{area legend,fill=blue!60!white}
    \addlegendimage{area legend,fill=orange!80!white}
    \end{customlegend}
    \end{tikzpicture}
    \caption{Intrinsic dataset variance measured using squared geodesic distance to the dataset Fr\'echet mean across different geometries. From left to right: DINOv2 ViT-L/14 with Euclidean geometry, CLIP ViT-L/14 with spherical geometry, and HyCoCLIP-B with hyperbolic geometry. UT-Zappos exhibits substantially lower variance than the broader datasets, while geometry-specific whitening increases the variance of UTZ-W toward the same regime.}
    \label{fig:primitive_win}
\end{figure}
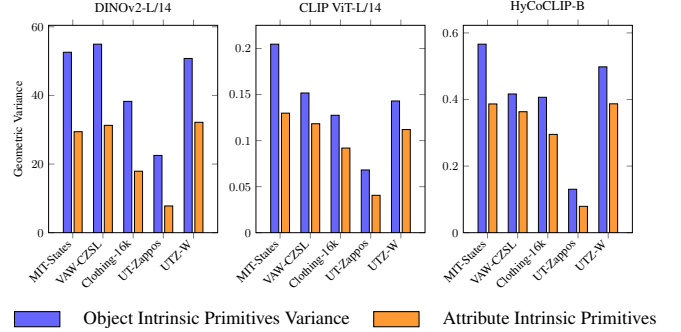

\noindent\textbf{Dataset structure affects compositional reasoning.}
UT-Zappos is a consistent exception to the advantages observed for primitive-based CZSL inference  on the broader CZSL datasets.
The dataset covers a narrow semantic domain and its samples have strongly coupled attribute--object relationship, with many attributes appearing across a restricted set of shoe categories (see Figure~\ref{fig:example_grid} in the Appendix).
We therefore test whether effectiveness of primitive-based compositional inference depends on the semantic spread of the embedding distribution: we investigate the intrinsic dataset variance and causally link its value to compositional generalization performance.
For embeddings $\{\mathbf{x}_i\}_{i=1}^{N}$ from dataset $\mathcal{D}$ on manifold $\mathcal{M}$, we compute the intrinsic variance
\begin{equation}
    V(\mathcal{D})
    =
    \frac{1}{N}
    \sum_{i=1}^{N}
    d_{\mathcal{M}}^2
    \left(
        \mathbf{x}_i,
        \boldsymbol{\mu}_{\mathcal{D}}
    \right),
\end{equation}
where $\boldsymbol{\mu}_{\mathcal{D}}$ is the Fr\'echet mean and $d_{\mathcal{M}}(\cdot,\cdot)$ is the geodesic distance.
Figure~\ref{fig:primitive_win} reveals a similar pattern across Euclidean, spherical, and hyperbolic representations: MIT-States, VAW-CZSL, and Clothing16K have substantially greater intrinsic variance than UT-Zappos.
This suggests that primitive decomposition benefits from representations spanning sufficiently diverse semantic contexts, whereas low-variance distributions allow pair representations to exploit stronger dataset-specific attribute-object co-occurrence patterns.

We study the role of intrinsic dataset variance as a causal factor to explain compositional reasoning limitations. We apply a geometry-specific whitening operation (detailed in Appendix~\ref{app:non_euclidean_whitening}) to the embeddings of the considered models of the UTZappos dataset, and denote the resulting representations as UTZ-W.
Whitening has the effect of increasing the intrinsic variance of UTZ-W (similarly to that of the broader datasets) compared to that of the original UT-Zappos. 

We report CZSL results before and after whitening in Table~\ref{tab:ut_zap}: the HM and UPA values increase noticeably for both GDE- and intrinsic-primitive compositional inference, achieving comparable or largely superior performance in some cases than pair-based inference. 

After whitening, intrinsic primitives achieve the highest HM for every model and the highest UPA for four of five models, falling only 0.05 points below the best MERU-B UPA results using pair-based inference. These results provide causal evidence that low intrinsic variance of the dataset (and its embedding distribution) contributes to the original failure mode of compositional inference. UT-Zappos is thus not a failure of primitive inference: its low-variance representation suppresses decomposition effectiveness.

\subsection{Statistical Separation of GDE Primitives}
\label{sec:factorization}
\begin{table*}[t]
\centering
\caption{\textbf{Group-robustness results (\%).} Average Accuracy (Avg.\ Acc.), Worst-Group Accuracy (WG.\ Acc.), and their gap on Waterbirds and CelebA. GDE and intrinsic object primitives consistently improve over zero-shot object inference without retraining or group annotations, with especially large gains for the hyperbolic models MERU and HyCoCLIP.}
\vspace{-1mm}
\label{tab:robustness}
{
    \fontsize{8 pt}{10 pt}\selectfont
    \begin{tabular}{@{}ll ccc ccc ccc@{}}
\toprule
& 
& \multicolumn{3}{c}{\textbf{Zero Shot}}
& \multicolumn{3}{c}{\textbf{GDE-Primitive}}
& \multicolumn{3}{c}{\textbf{Intrinsic Primitive}} \\
\cmidrule(lr){3-5}
\cmidrule(lr){6-8}
\cmidrule(lr){9-11}
& \textbf{Model}
& {Avg. Acc. $\uparrow$} & {WG. Acc. $\uparrow$} & {Gap $\downarrow$}
& {Avg. Acc. $\uparrow$} & {WG. Acc. $\uparrow$} & {Gap $\downarrow$}
& {Avg. Acc. $\uparrow$} & {WG. Acc. $\uparrow$} & {Gap $\downarrow$} \\
\midrule

\multirow{5}{*}{\makebox[0.5cm]{\rotatebox[origin=c]{90}{Waterbirds}}}
& CLIP ViT-L/14
& 80.9 & 48.8 & 32.1
& \textbf{90.8} & \textbf{85.7} & \textbf{5.13} 
& \textbf{90.8} & \textbf{85.7} & 5.17 \\

& DINOv2 ViT-L/14
& 69.1 & 42.4 & \textbf{26.7}
& 85.5 & 48.0 & 37.5
& \textbf{85.0} & \textbf{49.4} & 35.6 \\

& CLIP-B (GRIT)
& 58.4 & 36.9 & 21.5
& \textbf{77.1} & \textbf{64.4} & 12.7
& 77.0 & \textbf{64.4} & \textbf{12.7} \\

& MERU-B
& 58.3 & 31.7 & 26.6
& 63.0 & 39.0 & 24.0
& \textbf{77.1} & \textbf{62.6} & \textbf{14.6} \\

& HyCoCLIP-B
& 53.9 & 20.3 & 33.7
& 62.6 & 38.3 & 24.3
& \textbf{77.1} & \textbf{62.6} & \textbf{14.5} \\

\midrule

\multirow{5}{*}{\makebox[0.5cm]{\rotatebox[origin=c]{90}{CelebA}}}
& CLIP ViT-L/14
& 75.3 & 69.2 & 6.16
& \textbf{87.9} & \textbf{86.9} & \textbf{1.00} 
& 87.8 & 86.8 & \textbf{1.00} \\

& DINOv2 ViT-L/14
& 13.7 & 0.07 & 13.6
& \textbf{77.7} & 73.1 & 4.54 
& 77.4 & \textbf{73.8} & \textbf{3.57} \\

& CLIP-B (GRIT)
& 56.6 & 29.8 & 26.9
& \textbf{81.9} & \textbf{74.8} & 7.08 
& 81.7 & 74.7 & \textbf{6.97} \\

& MERU-B
& 35.2 & 11.5 & 23.6
& 69.1 & 59.6 & 9.58
& \textbf{84.1} & \textbf{80.3} & \textbf{3.80} \\

& HyCoCLIP-B
& 34.3 & 6.46 & 27.8
& 69.4 & 57.3 & 12.1
& \textbf{84.6} & \textbf{79.7} & \textbf{4.95} \\

\bottomrule
\end{tabular}%
}
\vspace{-3mm}
\end{table*}
A key premise of primitive-based inference is that the decomposition isolates object and attribute information into distinct semantic components.
We test this by measuring the statistical dependence between object, attribute, and pair representations using Pearson cross-correlation (CORR) and mutual information (MI). We perform this analysis on the GDE primitives.
The results show that, across datasets and model families, the dependence and mutual information between object and attribute primitive representations  is consistently substantially lower than those of either object and pair or attribute and pair representations.
For example, on MIT-States object and attribute primitive CORR remains between $0.444$ and $0.473$ across models, compared to $0.825$--$0.895$ for object primitive and pair representation, and the MI estimates are $13.0$ versus $119.4$--$320.7$.
We observe similar results across the considered CZSL datasets and model representations. 
We report full results in Appendix~\ref{app:corr_mi}.

The higher dependence that we consistently observe between the primitive representations and the full pair representation are expected and confirm the disentanglement quality of the factorization that we perform. Object and attribute primitives retain information from the composition from which they are derived, while exhibiting weak dependence on one another.
These results provide evidence that primitive-based decomposition, as we study it, recovers a factorized semantic structure in which object and attribute information is meaningfully separated rather than simply reproducing the original pair representation, and thus providing a solid tool for compositional reasoning.

\subsection{Group Robustness via Primitives}

Pre-trained vision-language models achieve strong zero-shot performance under distribution shift, but remain sensitive to spurious correlations and subgroup imbalance~\cite{zhang2022contrastive}. Existing approaches to group robustness typically rely on group annotations, additional optimization, group-specific prompts, adaptation, or class reweighting~\cite{sagawaDistributionallyRobustNeural2019,Sohoni20,kim2025groupcoopgrouprobustfinetuninggroup,gourabathina2026robustnessknowngroupslowrank,kirichenko2023last,Seo2025}. 

We instead ask whether the compositional structure already encoded in and emerging from pretrained representations can itself be used for robust inference. 
The results of the statistical disentanglement indicate that decomposed primitives separate object and attribute semantics, allowing inference to retain information relevant to the object while suppressing attribute directions that may induce spurious correlations. We thus evaluate decomposed primitive for robustness to group bias on Waterbirds~\cite{sagawaDistributionallyRobustNeural2019} and CelebA~\cite{liuDeepLearningFace2015}. We do not use group annotations, additional data, or further optimization. Specifically, we classify images using only the object primitives, retaining only the object-score component of Eq.~\ref{eq:zs-prim}, Eq.~\ref{eq:intr-prim} and Eq.~\ref{eq:gde-prim} for zero-shot, intrinsic, and GDE primitives, respectively.
Table~\ref{tab:robustness} shows that this simple inference-time decomposition consistently improves group robustness. Both GDE and intrinsic object primitives improve worst-group accuracy over zero-shot object inference while largely preserving, and in several cases improving, average accuracy. These gains arise without an explicit debiasing objective or fairness supervision: they come solely from isolating semantic structure already present in the pretrained representations. This suggests that compositional factorization is not only useful for interpreting representations, but also provides a practical mechanism for reducing reliance on spurious attributes at inference time.

\noindent\textbf{Intrinsic primitives yield better group robustness in hyperbolic spaces.}
The group robustness gains are especially pronounced for MERU and HyCoCLIP, where intrinsic object primitives yield large improvements in both average and worst-group accuracy. We conjecture that this advantage is related to the radial organization of semantic abstraction in the hyperbolic space. Intrinsic object primitives lie closer to the origin than compound object--attribute representations, consistent with their encoding broader, less attribute-specific concepts. GDE primitives, by contrast, are obtained through a local linear approximation around the Fr\'{e}chet mean and can therefore remain more influenced by the local neighborhood structure induced by dominant attribute or background correlations. This provides a plausible geometric explanation for the stronger robustness of intrinsic primitives that lay directly on the manifold. The radial organization of these representations is further explored in Appendix~\ref{app:hierarchy_hyperbolic}.

\section{Conclusion}\label{sec:conc}
This work establishes primitive decomposition as a unifying principle for compositional zero-shot learning (CZSL) across Euclidean, spherical, and hyperbolic vision–language models. By systematically analyzing models with diverse manifold geometries, we demonstrate that compositional structure consistently emerges in pretrained vision embeddings in the form of primitive visual concept representations, which can be directly exploited for compositional zero-shot inference. Crucially, we show that explicit attribute-object pair re-composition is not necessary for CZSL: independently reasoning over attribute and object primitives, with specifically-designed scoring functions, can effectively support compositional open-world generalization.

We demonstrate primitives as a novel paradigm for CZSL inference with results being consistent for the tangent space GDE primitives and intrinsic primitives.
Across experiments, GDE primitives provide robust and transferable primitive representations, consistently outperforming pair-based compositions for open-world CZSL while remaining simple and geometry-agnostic. Intrinsic primitives offer a complementary, geometry-aware refinement, particularly effective in hyperbolic spaces, where operating directly on the manifold improves stability by avoiding local linearization errors.
We further demonstrate benefits of primitive-based inference in group robustness tasks, where object-primitive representations allow to perform unbiased inference without model re-training, avoiding spurious correlations induced by attribute characteristics.

Finally, this work provides the first comprehensive and unified view of primitive-based CZSL, demonstrating it as an emerging property of existing pretrained models.

\section*{Acknowledgments}
\noindent This publication is part of the project ROBUST: Trustworthy AI-based Systems for Sustainable Growth with project number KICH3.LTP.20.006, which is (partly) financed by the Dutch Research Council (NWO), Siemens Healthineers, and the Dutch Ministry of Economic Affairs and Climate Policy (EZK) under the program LTP KIC 2020--2023.

\noindent This publication is part of the project ReVision with file number  VI.Vidi.243.176 of the research programme Vidi ENW which is (partly) financed by the Dutch Research Council (NWO) under the grant \url{ https://doi.org/10.61686/ERNED50610}.

\noindent T.T. acknowledges the EU project ELSA-European Lighthouse on Secure and Safe AI, under the EU’s Horizon Europe Research and Innovation Programme (grant agreement no. 101070617)

\noindent This publication is part of the project HyperVision with file number VI.Vidi.243.085 of the research programme Vidi ENW which is (partly) financed by the Dutch Research Council (NWO) under the grant \url{https://doi.org/10.61686/DCECG94893}.

{
    \small
    \bibliographystyle{ieeenat_fullname}
    \bibliography{references}
}

\clearpage

\appendix

\section{Riemannian Gradient Descent}
\label{app:intrinsic_mean_calc}

Given a collection of manifold-valued embeddings $\{f(x_i)\}_{i=1}^N \subset \mathcal{M}$,
we seek a representative point that minimizes the sum of squared geodesic distances to all samples. This point is known as the \emph{intrinsic mean}  (or Fréchet mean) and is defined as
\[
\mu^\star
=
\arg\min_{\mu \in \mathcal{M}}
\frac{1}{N}
\sum_{i=1}^N
d_{\mathcal{M}}^2\!\left(\mu, f(x_i)\right),
\]
where \(d_{\mathcal{M}}(\cdot,\cdot)\) denotes the geodesic distance on the
manifold \(\mathcal{M}\).
Unlike Euclidean averaging, optimization on a curved manifold requires moving
along tangent spaces using the logarithmic and exponential maps. At an iterate
\(\mu^{(t)} \in \mathcal{M}\), each sample \(f(x_i)\) is first mapped to the
tangent space \(T_{\mu^{(t)}}\mathcal{M}\) via the logarithmic map $\mathrm{Log}_{\mu^{(t)}}(f(x_i))$. The average tangent vector:
\[
\mathbf{v}^{(t)}
=
\frac{1}{N}
\sum_{i=1}^N
\mathrm{Log}_{\mu^{(t)}}(f(x_i))
\]
corresponds to the Riemannian gradient direction of the Fréchet objective. The iterate is then updated by mapping this tangent vector back onto the manifold using the exponential map:
\[
\mu^{(t+1)}
=
\mathrm{Exp}_{\mu^{(t)}}\!\left(
\eta \mathbf{v}^{(t)}
\right),
\]
where \(\eta > 0\) is the step size.
Algorithm~\ref{alg:frechet_mean} summarizes the complete optimization
procedure.

\begin{algorithm}[h]
\caption{Intrinsic Mean Computation}
\label{alg:frechet_mean}
\begin{algorithmic}[1]
\Require Embeddings $\{f(x_i)\}_{i=1}^N \subset \mathcal{M}$,
initial point $\boldsymbol{\mu}^{(0)} \in \mathcal{M}$,
step size $\eta$
\For{$t=0,1,\dots,T-1$}
    \State
    $\mathbf{v}^{(t)}
    \gets
    \frac{1}{N}
    \sum_{i=1}^N
    \mathrm{Log}_{\boldsymbol{\mu}^{(t)}}(f(x_i))$
    \State
    $\boldsymbol{\mu}^{(t+1)}
    \gets
    \mathrm{Exp}_{\boldsymbol{\mu}^{(t)}}(
    \eta \mathbf{v}^{(t)})$
\EndFor
\State \Return $\boldsymbol{\mu}^{(T)}$
\end{algorithmic}
\end{algorithm}

\section{Dense Feature Extraction via Talk2DINO}
\label{app:dino_dense}

Unlike CLIP, MERU, or HyCoCLIP that inherently produce aligned image-text features from joint text pairs, vision-only models like DINOv2 yield dense patch embeddings without text alignment. To construct a competitive text-conditioned compositional baseline for DINOv2 that matches the open-world zero-shot capabilities of vision-language models, we interface its Euclidean representation space via Talk2DINO~\cite{barsellottiTalkingDINOBridging2024}.
Given an image $x$ and a textual query $q$, Talk2DINO produces patch embeddings
$\{\mathbf{p}_m\}_{m=1}^{M}$ together with a query embedding $\mathbf{q}$.
The spatial relevance map and the final pooled image representation are computed respectively 
as
\begin{align}
    \alpha_m
    =
    \frac{
        \exp\!\left(
            \tau
            \frac{
                \mathbf{p}_m^\top\mathbf{q}
            }{
                \|\mathbf{p}_m\|\,\|\mathbf{q}\|
            }
        \right)
    }{
        \sum_{k=1}^{M}
        \exp\!\left(
            \tau
            \frac{
                \mathbf{p}_k^\top\mathbf{q}
            }{
                \|\mathbf{p}_k\|\,\|\mathbf{q}\|
            }
        \right)
    },
    \\
    f(\{p_m\}_{m=1}^M,q)
    =
    \sum_{m=1}^{M}\alpha_m\mathbf{p}_m,
\end{align}
where $\tau$ is a temperature parameter. This routing mechanism yields a single, unified Euclidean image representation conditioned dynamically on the queried attribute-object composition while strictly preserving DINOv2's original underlying geometry.

\section{DINO Embeddings: Euclidean or Spherical?}
\label{app:abl_dino_euc_sph}
Talk2DINO provides a language wrapper to DINO, without modifying its native visual embeddings. This enables us to compute for DINO the for fair comparison. 
The core compositionality analysis is performed on the native DINO vision embedding and confirms that properly addressing manifold geometry is key to effective decomposition:  Table~\ref{tab:dinov2_euc_sph} shows that 
reducing the embeddings to spherical space harms performance, due to wrong manifold assumption for primitive extraction. 
\begin{table}[h]
    \centering
    \resizebox{\linewidth}{!}{
    \begin{tabular}{@{}ll ccc ccc@{}}
    \toprule
    & & \multicolumn{3}{c}{\textbf{Harmonic Mean (\%, $\uparrow$)}} & \multicolumn{3}{c}{\textbf{Unseen Pair Accuracy (\%, $\uparrow$)}} \\
    \cmidrule(lr){3-5} \cmidrule(lr){6-8}
    \textbf{Dataset} & $\mathcal{M}$ & \textbf{GDE Pair} & \textbf{GDE Prim.} & \textbf{Int. Prim.} & \textbf{GDE Pair} & \textbf{GDE Prim.} & \textbf{Int. Prim.} \\
    \midrule

    \multirow{2}{*}{\makebox[0.5cm]{\rotatebox[origin=c]{90}{MIT}}} 
    & $\mathrm{euc}$ & 6.56 & 8.22 & 8.04 & 6.26 & 9.33 & 9.42 \\
    & $\mathrm{sph}$ & 2.17 & 4.70 & 4.85 & 1.87 & 4.61 & 4.75 \\
    \midrule

    \multirow{2}{*}{\makebox[0.5cm]{\rotatebox[origin=c]{90}{VAW}}} 
    & $\mathrm{euc}$ & 1.30 & 2.05 & 2.20 & 1.63 & 2.60 & 2.81 \\
    & $\mathrm{sph}$ & 0.28 & 0.68 & 0.72 & 0.28 & 0.65 & 0.71 \\
    
    \bottomrule
    \end{tabular}
    }\vspace{-3mm}
    \caption{
    Evaluation results utilizing the Talk2DINO as Euclidean ($\mathrm{euc}$) or Spherical ($\mathrm{sph}$) embeddings (after $\ell_2$ normalisation). The native euclidean manifold substantially outperforms spherical. 
    }
    \label{tab:dinov2_euc_sph}
\end{table}

\section{GDE Primitive Concept Separation}
\label{app:corr_mi}

To assess whether the GDE primitives capture concept structure that is separable in the model's embedding space, we measure the statistical dependence between each pair of primitives using two complementary estimators: the Pearson cross correlation (CORR), and a canonical correlation estimator of mutual information (MI) computed with the \texttt{bench-mi} mutual information package~\cite{Czyz2024Sep}.
CORR captures linear alignment between two sets of vectors, while MI additionally captures non-linear dependence structure that a purely linear estimator would miss. 
We report both for the three primitive pairings: object-attribute, object-pair, and attribute-pair.
Our analysis is limited to GDE primitives.

\begin{table*}[t]
\centering
\caption{Cross Correlation and Mutual Information estimates between the GDE primitives and the model embedding. Since GDE primtives live in euclidean space, usual statistics calculation methods apply.}
\label{tab:mutual_information}
{
    \fontsize{8 pt}{10 pt}\selectfont
    \begin{tabular}{llcccccccccc}
\toprule
 &  & \multicolumn{2}{c}{CLIP ViT-L/14} & \multicolumn{2}{c}{DINOv2 ViT-L/14} & \multicolumn{2}{c}{CLIP-B (GRIT)} & \multicolumn{2}{c}{HyCoCLIP-B} & \multicolumn{2}{c}{MERU-B} \\
\cmidrule(lr){3-4}\cmidrule(lr){5-6}\cmidrule(lr){7-8}\cmidrule(lr){9-10}\cmidrule(lr){11-12}
 &  & CORR & MI & CORR & MI & CORR & MI & CORR & MI & CORR & MI \\
\midrule
\multirow{3}{*}{MIT-States} & object-attribute & 0.453 & 13.0 & 0.456 & 13.0 & 0.444 & 13.0 & 0.453 & 13.0 & 0.473 & 13.0 \\
 & object-pair & 0.825 & 191.0 & 0.895 & 320.7 & 0.831 & 148.7 & 0.839 & 119.7 & 0.843 & 119.4 \\
 & attribute-pair & 0.530 & 77.9 & 0.516 & 129.0 & 0.511 & 44.5 & 0.513 & 44.2 & 0.531 & 44.3 \\
\midrule
\multirow{3}{*}{VAW-CZSL} & object-attribute & 0.187 & 7.2 & 0.183 & 7.2 & 0.195 & 6.9 & 0.191 & 6.9 & 0.173 & 6.9 \\
 & object-pair & 0.575 & 38.1 & 0.713 & 61.0 & 0.565 & 28.1 & 0.565 & 26.6 & 0.576 & 29.9 \\
 & attribute-pair & 0.203 & 14.9 & 0.228 & 13.2 & 0.215 & 16.5 & 0.213 & 11.0 & 0.203 & 10.9 \\
\midrule
\multirow{3}{*}{UT-Zappos} & object-attribute & 0.098 & 0.8 & 0.090 & 0.8 & 0.067 & 0.8 & 0.067 & 0.8 & 0.069 & 0.8 \\
 & object-pair & 0.872 & 4.8 & 0.930 & 5.1 & 0.896 & 5.5 & 0.909 & 5.6 & 0.929 & 18.0 \\
 & attribute-pair & 0.338 & 4.0 & 0.174 & 3.4 & 0.264 & 3.0 & 0.241 & 3.0 & 0.208 & 3.1 \\
\midrule
\multirow{3}{*}{Clothing16k} & object-attribute & 0.404 & 2.7 & 0.454 & 2.7 & 0.419 & 2.7 & 0.427 & 2.7 & 0.419 & 2.7 \\
 & object-pair & 0.945 & 73.1 & 0.992 & 73.0 & 0.962 & 72.7 & 0.953 & 73.2 & 0.948 & 73.0 \\
 & attribute-pair & 0.606 & 83.1 & 0.488 & 83.2 & 0.581 & 83.4 & 0.612 & 83.7 & 0.609 & 83.2 \\
\bottomrule
\end{tabular}

}
\end{table*}
Across datasets and models, in Table~\ref{tab:mutual_information}, we observe trends consistent with our expectations: the object-attribute pairing is consistently the lowest correlated and carries the least mutual information relative to object-pair and attribute-pair. 
This suggests that object and attribute representations are comparatively more disentangled from one another, whereas information about objects and attributes is each more entangled with pairwise (relational) structure than with each other.

We also note that object-pair estimates are consistently higher than attribute-pair estimates across datasets and models. 
This points to embeddings being generally more aligned with object-level information than with attribute-level information, and suggests attribute-awareness may be an area where current embeddings are comparatively weaker, a direction that could benefit from further study. 

\section{Non-Euclidean Whitening}
\label{app:non_euclidean_whitening}

For Euclidean features, whitening can be applied directly in the ambient space.
For spherical and hyperbolic features, we instead perform whitening in a tangent space, following the general principle of Principal Geodesic Analysis (PGA) \cite{pga}.
The intrinsic mean $\boldsymbol{\mu}$ is computed as described in Algorithm~\ref{alg:frechet_mean}, after which the embeddings are mapped to the corresponding tangent space using the logarithmic map.

\subsection{Spherical Whitening}

For spherical embeddings $\{f(x_i)\}_{i=1}^N \subset \mathbb{S}^{d-1}$, we first compute the intrinsic mean $\boldsymbol{\mu}$ and map each embedding to $T_{\boldsymbol{\mu}}\mathbb{S}^{d-1}$ using $\mathrm{Log}_{\boldsymbol{\mu}}$, as defined in Eq.~\ref{eq:sphere_maps}.
Whitening is then performed on the resulting tangent vectors.

A spherical-specific restriction arises from the cut locus.
For the unit sphere, the logarithmic representation around $\boldsymbol{\mu}$ is unique only for tangent vectors with norm smaller than $\pi$.
Arbitrarily rescaling a tangent vector beyond this radius causes the exponential map to wrap around the sphere.
Distinct tangent vectors can then map to the same or nearby spherical locations, which may destroy semantic information encoded by their relative positions. We therefore constrain the norm of the projected tangent vector to remain within the injectivity radius with rescaling. This follows the same general tangent-space construction used in spherical PCA methods \texttt{spherepc} \cite{spherpc}.

\begin{algorithm}[t]
\caption{Spherical whitening}
\label{alg:spherical_whitening}
\begin{algorithmic}[1]
\Require Embeddings $\{f(x_i)\}_{i=1}^N \subset \mathbb{S}^{d-1}$
\State Compute intrinsic mean $\boldsymbol{\mu}$ using Algorithm~\ref{alg:frechet_mean}
\For{$i=1,\dots,N$}
    \State $\mathbf{v}_i \gets \mathrm{Log}_{\boldsymbol{\mu}}(f(x_i))$
\EndFor
\State Compute covariance $\mathbf{C}$ of $\{\mathbf{v}_i\}_{i=1}^N$
\State Compute whitening transform $\mathbf{W} \gets \mathbf{C}^{-1/2}$
\For{$i=1,\dots,N$}
    \State $\hat{\mathbf{v}}_i \gets \mathbf{W}\mathbf{v}_i$
\EndFor
\State Rescale $\{\hat{\mathbf{v}}_i\}_{i=1}^N$ if required such that $\|\hat{\mathbf{v}}_i\| < \pi$
\For{$i=1,\dots,N$}
    \State $\hat{f}(x_i) \gets \mathrm{Exp}_{\boldsymbol{\mu}}(\hat{\mathbf{v}}_i)$
\EndFor
\State \Return $\{\hat{f}(x_i)\}_{i=1}^N$
\end{algorithmic}
\end{algorithm}

\subsection{Hyperboloid Whitening}

For hyperbolic embeddings $\{f(x_i)\}_{i=1}^N \subset \mathbb{H}^{d+1}$, we again begin by computing an intrinsic mean $\boldsymbol{\mu}$.
Unlike the spherical case, the ambient hyperboloid coordinates are equipped with the Lorentz metric rather than the Euclidean metric.
We therefore first apply a Lorentz isometry that maps $\boldsymbol{\mu}$ to the canonical hyperbolic origin $\mathbf{o}$.
This preserves hyperbolic distances while placing the tangent representation in a canonical coordinate system.
Related use of hyperbolic isometries appears in HoroPCA \cite{horopca}.

After this transformation, the embeddings are mapped to $T_{\mathbf{o}}\mathbb{H}^{d+1}$ using the logarithmic map as defined in Eq.~\ref{eq:hyp_maps}.
At the canonical origin, the tangent coordinates provide an ordinary Euclidean representation of the local Riemannian metric, and standard covariance whitening can therefore be applied.
The whitened tangent vectors are finally mapped back to the hyperboloid using the exponential map.

In contrast to the sphere, hyperbolic space has no finite cut locus.
The tangent radius is therefore not subject to the same wrapping ambiguity.

\begin{algorithm}[t]
\caption{Hyperboloid whitening}
\label{alg:hyperboloid_whitening}
\begin{algorithmic}[1]
\Require Embeddings $\{f(x_i)\}_{i=1}^N \subset \mathbb{H}^{d+1}$
\State Compute intrinsic mean $\boldsymbol{\mu}$ using Algorithm~\ref{alg:frechet_mean}
\State Compute Lorentz isometry $\mathcal{B}_{\boldsymbol{\mu}}$ such that $\mathcal{B}_{\boldsymbol{\mu}}(\boldsymbol{\mu})=\mathbf{o}$
\For{$i=1,\dots,N$}
    \State $\mathbf{u}_i \gets \mathcal{B}_{\boldsymbol{\mu}}(f(x_i))$
    \State $\mathbf{v}_i \gets \mathrm{Log}_{\mathbf{o}}(\mathbf{u}_i)$
\EndFor
\State Compute covariance $\mathbf{C}$ of $\{\mathbf{v}_i\}_{i=1}^N$
\State Compute whitening transform $\mathbf{W} \gets \mathbf{C}^{-1/2}$
\For{$i=1,\dots,N$}
    \State $\hat{\mathbf{v}}_i \gets \mathbf{W}\mathbf{v}_i$
    \State $\hat{f}(x_i) \gets \mathrm{Exp}_{\mathbf{o}}(\hat{\mathbf{v}}_i)$
\EndFor
\State \Return $\{\hat{f}(x_i)\}_{i=1}^N$
\end{algorithmic}
\end{algorithm}

\section{Primitive Weighting Ablation}
\label{app:abl_weighing}
GDE primitive sample weights are scalar, carrying only confidence rather than semantic information. 
In Table~\ref{tab:noise_ablation}, we provide additional results that ablate this design inherited  from the baseline framework.
\begin{table}[t]
    \centering
    \resizebox{\linewidth}{!}{
    \begin{tabular}{@{}ll cccc@{}}
    \toprule
    & & \multicolumn{2}{c}{\textbf{Harmonic Mean (\%, $\uparrow$)}} & \multicolumn{2}{c}{\textbf{Unseen Pair Accuracy (\%, $\uparrow$)}} \\
    \cmidrule(lr){3-4} \cmidrule(lr){5-6}
    \textbf{Dataset} & \textbf{Model} & \textbf{Weighted} & \textbf{Uniform} & \textbf{Weighted} & \textbf{Uniform} \\
    \midrule

    \multirow{5}{*}{\makebox[0.5cm]{\rotatebox[origin=c]{90}{MIT}}}
    & CLIP ViT-L/14    & 10.90 & 10.39 & 11.98 & 11.47 \\
    & DINOv2 ViT-L/14  & 8.22  & 8.02  & 9.33  & 9.10  \\
    & CLIP-B (GRIT)    & 6.64  & 6.12  & 7.18  & 6.62  \\
    & MERU-B           & 6.57  & 6.59  & 7.25  & 7.14  \\
    & HyCoCLIP-B       & 7.94  & 7.83  & 8.75  & 8.62  \\
    \midrule

    \multirow{5}{*}{\makebox[0.5cm]{\rotatebox[origin=c]{90}{VAW}}}
    & CLIP ViT-L/14    & 1.77  & 1.64  & 2.07  & 2.03  \\
    & DINOv2 ViT-L/14  & 2.05  & 2.11  & 2.60  & 2.64  \\
    & CLIP-B (GRIT)    & 1.52  & 1.37  & 2.13  & 2.01  \\
    & MERU-B           & 1.49  & 1.43  & 1.97  & 1.87  \\
    & HyCoCLIP-B       & 1.47  & 1.41  & 2.09  & 2.03  \\
    \bottomrule
    \end{tabular}
    }
    \caption{
   Disabling noise estimation results in minor performance variations across models for GDE primitives validating that weighing carries no directional information.
    }
    \label{tab:noise_ablation}
\end{table}

\section{Extended Open-World CZSL Metrics}
\label{app:extended_metrics}

This section reports additional evaluation metrics for open-world compositional
zero-shot learning beyond the harmonic mean discussed in the main paper.
In addition to composed pair performance, we separately evaluate
object-level, attribute-level.
These metrics provide a more detailed view of compositional behaviour.
Object and attribute accuracies isolate recognition performance on the
underlying semantic primitives, helping distinguish whether improvements arise
from better primitive identification or from stronger compositional reasoning.

Table~\ref{tab:obj_attr} reports object and attribute accuracies across all
datasets and geometries.
Consistent with the main results, primitive-based inference generally improves
both object and attribute recognition, particularly on datasets with greater
semantic diversity such as MIT-States and VAW-CZSL.
These gains indicate that primitive decomposition improves not only pair
retrieval, but also the separability of the underlying semantic factors.
Table~\ref{tab:seen_unseen} reports seen and overall pair accuracies.

\begin{table*}[t]
\centering
\caption{%
  \textbf{Open-world Compositional Zero-Shot Learning results (\%, $\uparrow$).}
  Metrics include Object Accuracy and Attribute Accuracy.
}
\label{tab:obj_attr}

\resizebox{\linewidth}{!}{%
    \begin{tabular}{
@{}ll 
ccccc 
ccccc@{}
}
\toprule

& &
\multicolumn{5}{c}{\textbf{Object Accuracy (\%, $\uparrow$)}} &
\multicolumn{5}{c}{\textbf{Attribute Accuracy (\%, $\uparrow$)}} \\

\cmidrule(lr){3-7}
\cmidrule(lr){8-12}

 &
\textbf{Model} &
\textbf{ZS Pair} &
\textbf{ZS Prim.} &
\textbf{GDE Pair} &
\textbf{GDE Prim.} &
\textbf{Int. Prim.} &
\textbf{ZS Pair} &
\textbf{ZS Prim.} &
\textbf{GDE Pair} &
\textbf{GDE Prim.} &
\textbf{Int. Prim.} \\

\midrule

\multirow{5}{*}{\makebox[0.5cm]{\rotatebox[origin=c]{90}{MIT-States}}}
& CLIP ViT-L/14    & 47.33 & 47.33 & 43.79 & 43.68 & \textbf{43.94} 
                   & 15.76 & 20.05 & 18.38 & 20.57 & \textbf{20.68} \\
& DINOv2 ViT-L/14  & 32.70 & 29.70 & 40.13 & \textbf{40.25} & 40.22 
                   & 8.49 & 11.97 & 16.00 & \textbf{18.45} & 18.41 \\
& CLIP-B (GRIT)    & \textbf{36.59} & 35.81 & 33.91 & 35.50 & 35.58 
                   & 10.35 & 13.48 & 14.09 & \textbf{17.41} & 16.97 \\
& MERU-B           & \textbf{37.31} & 36.34 & 34.03 & 35.39 & 35.51 
                   & 10.57 & 13.22 & 11.71 & \textbf{17.48} & 17.07 \\
& HyCoCLIP-B       & \textbf{39.79} & 39.72 & 37.41 & 39.01 & 39.15 
                   & 10.95 & 16.26 & 12.00 & 18.09 & 17.71 \\

\midrule

\multirow{5}{*}{\makebox[0.5cm]{\rotatebox[origin=c]{90}{VAW-CZSL}}}
& CLIP ViT-L/14    & 18.66 & 19.18 & 24.49 & \textbf{25.99} & 25.33 
                   & 3.70 & 5.97 & 5.40 & \textbf{6.69} & 6.37 \\
& DINOv2 ViT-L/14  & 20.77 & 20.80 & 30.07 & \textbf{30.88} & \textbf{30.88} 
                   & 4.24 & 5.38 & 5.82 & \textbf{8.06} & 7.85 \\
& CLIP-B (GRIT)    & 13.94 & 15.58 & 20.22 & \textbf{22.29} & 21.78 
                   & 3.19 & 4.39 & 5.55 & \textbf{6.72} & 6.32 \\
& MERU-B           & 14.54 & 15.60 & 20.91 & \textbf{22.41} & 22.13 
                   & 2.92 & 4.38 & 5.20 & \textbf{6.80} & 6.03 \\
& HyCoCLIP-B       & 15.96 & 19.12 & 21.77 & \textbf{22.96} & 22.61 
                   & 2.64 & 4.75 & 4.81 & \textbf{6.45} & 6.19 \\

\midrule

\multirow{5}{*}{\makebox[0.5cm]{\rotatebox[origin=c]{90}{UT-Zappos}}}
& CLIP ViT-L/14    & 50.03 & \textbf{63.90} & 60.33 & 57.93 & 59.75 
                   & 18.53 & 17.54 & \textbf{25.91} & 19.73 & 21.69 \\
& DINOv2 ViT-L/14  & 33.22 & 12.32 & 56.04 & 55.11 & \textbf{56.21} 
                   & 13.90 & 13.25 & \textbf{25.91} & 6.35 & 5.66 \\
& CLIP-B (GRIT)    & 36.65 & 29.89 & 49.83 & 47.39 & \textbf{50.62} 
                   & 11.77 & 10.16 & 18.50 & \textbf{23.58} & 22.10 \\
& MERU-B           & 21.86 & 17.19 & 50.48 & \textbf{51.41} & 50.27 
                   & 13.62 & 16.64 & 18.91 & 21.45 & \textbf{22.10} \\
& HyCoCLIP-B       & 34.45 & 25.57 & 46.88 & \textbf{48.70} & 47.67 
                   & 12.94 & 21.79 & 16.09 & 15.41 & \textbf{15.61} \\

\midrule

\multirow{5}{*}{\makebox[0.5cm]{\rotatebox[origin=c]{90}{Clothing16k}}}
& CLIP ViT-L/14    & 97.10 & 93.06 & 84.50 & \textbf{97.48} & 97.39 
                   & 95.11 & 93.29 & 64.28 & \textbf{95.17} & 95.11 \\
& DINOv2 ViT-L/14  & 78.76 & 65.63 & 84.32 & \textbf{79.81} & 78.99 
                   & 22.12 & \textbf{36.89} & 11.34 & 32.35 & 32.20 \\
& CLIP-B (GRIT)    & 93.11 & 92.82 & 87.72 & \textbf{94.02} & 93.82 
                   & 93.26 & 85.56 & 71.90 & \textbf{95.46} & 95.22 \\
& MERU-B           & 93.14 & 93.55 & 88.31 & 94.20 & \textbf{94.37} 
                   & 91.71 & 84.09 & 76.65 & \textbf{94.35} & 93.88 \\
& HyCoCLIP-B       & 92.73 & 92.88 & 89.25 & 94.08 & \textbf{94.55} 
                   & 93.20 & 94.32 & 78.23 & \textbf{95.43} & 94.52 \\

\bottomrule
\end{tabular}%
}
\end{table*}

\begin{table*}[t]
\centering
\caption{%
  \textbf{Open-world Compositional Zero-Shot Learning results (\%, $\uparrow$).}
  Metrics include Seen Pair Accuracy and Overall Pair Accuracy.
}
\label{tab:seen_unseen}

\resizebox{\linewidth}{!}{%
    \begin{tabular}{
@{}ll
ccccc
ccccc@{}
}
\toprule

& &
\multicolumn{5}{c}{\textbf{Seen Pair Accuracy (\%, $\uparrow$)}} &
\multicolumn{5}{c}{\textbf{Overall Pair Accuracy (\%, $\uparrow$)}} \\
\cmidrule(lr){3-7}
\cmidrule(lr){8-12}
 &
\textbf{Model} &
\textbf{ZS Pair} &
\textbf{ZS Prim.} &
\textbf{GDE Pair} &
\textbf{GDE Prim.} &
\textbf{Int. Prim.} &
\textbf{ZS Pair} &
\textbf{ZS Prim.} &
\textbf{GDE Pair} &
\textbf{GDE Prim.} &
\textbf{Int. Prim.} \\
\midrule
\multirow{5}{*}{\makebox[0.5cm]{\rotatebox[origin=c]{90}{MIT-States}}}
& CLIP ViT-L/14
    & 29.03 & 22.23 & \textbf{31.18} & 25.21 & 25.88
    & 15.76 & 20.05 & 18.38 & 20.57 & \textbf{20.68} \\
& DINOv2 ViT-L/14
    & 16.64 & 9.87 & \textbf{27.23} & 20.63 & 19.96
    & 8.49 & 11.97 & 16.00 & \textbf{18.45} & 18.41 \\
& CLIP-B (GRIT)
    & 21.72 & 16.93 & \textbf{22.18} & 18.99 & 18.82
    & 10.35 & 13.48 & 14.09 & \textbf{17.41} & 16.97 \\
& MERU-B
    & \textbf{22.61} & 16.76 & 19.58 & 18.91 & 19.20
    & 10.57 & 13.22 & 11.71 & \textbf{17.48} & 17.07 \\
& HyCoCLIP-B
    & \textbf{23.87} & 15.50 & 21.97 & 21.34 & 20.97
    & 10.95 & 16.26 & 12.00 & \textbf{18.09} & 17.71 \\
\midrule
\multirow{5}{*}{\makebox[0.5cm]{\rotatebox[origin=c]{90}{VAW-CZSL}}}
& CLIP ViT-L/14    & 3.76 & \textbf{4.97} & 3.67 & 4.87 & 4.67
                   & 3.70 & 5.97 & 5.40 & \textbf{6.69} & 6.37 \\
& DINOv2 ViT-L/14  & 3.64 & 4.21 & 3.62 & \textbf{5.44} & 5.14
                   & 4.24 & 5.38 & 5.82 & \textbf{8.06} & 7.85 \\
& CLIP-B (GRIT)    & 2.12 & 2.80 & 2.88 & 3.37 & \textbf{3.39}
                   & 3.19 & 4.39 & 5.55 &\textbf{ 6.72} & 6.32 \\
& MERU-B           & 2.31 & 3.22 & 2.49 & \textbf{3.79} & 3.44
                   & 2.92 & 4.38 & 5.20 & \textbf{6.80} & 6.03 \\
& HyCoCLIP-B       & 2.31 & \textbf{3.96} & 2.53 & 3.77 & 3.69
                   & 2.64 & 4.75 & 4.81 & \textbf{6.45} & 6.19 \\
\midrule
\multirow{5}{*}{\makebox[0.5cm]{\rotatebox[origin=c]{90}{UT-Zappos}}}
& CLIP ViT-L/14    & 13.49 & 8.80 & \textbf{29.03} & 19.16 & 21.70
                   & 18.53 & 17.54 & \textbf{25.91} & 19.73 & 21.69 \\
& DINOv2 ViT-L/14  & 2.25 & 0.29 & \textbf{18.48} & 10.36 & 7.82
                   & 13.90 & 13.25 & \textbf{25.91} & 6.35 & 5.66 \\
& CLIP-B (GRIT)    & 1.76 & 1.76 & 19.35 & 22.09 & \textbf{22.39}
                   & 11.77 & 10.16 & 18.50 & \textbf{23.58} & 22.10 \\
& MERU-B           & 3.23 & 4.11 & 21.41 & 22.78 & \textbf{23.07}
                   & 13.62 & 16.64 & 18.91 & 21.45 & \textbf{22.10} \\
& HyCoCLIP-B       & 9.48 & 3.71 & \textbf{17.69} & 11.24 & 13.88
                   & 12.94 & \textbf{21.79} & 16.09 & 15.41 & 15.61 \\
\midrule
\multirow{5}{*}{\makebox[0.5cm]{\rotatebox[origin=c]{90}{Clothing16k}}}
& CLIP ViT-L/14    & 98.72 & 91.30 & 87.21 & \textbf{98.98} & 98.72
                   & 92.41 & 86.90 & 58.86 & \textbf{92.76} & 92.62 \\
& DINOv2 ViT-L/14  & 48.59 & 42.46 & 52.94 & 52.17 & \textbf{57.54}
                   & 19.22 & 28.42 & 9.00 & \textbf{30.65} & 30.53 \\
& CLIP-B (GRIT)    & 97.19 & 90.03 & 92.07 & 97.19 & \textbf{97.44}
                   & 87.11 & 79.20 & 65.48 & \textbf{89.83} & 89.42 \\
& MERU-B           & 97.70 & 93.09 & 93.09 & \textbf{97.95} & 97.44
                   & 85.64 & 78.35 & 69.47 & \textbf{88.95} & 88.69 \\
& HyCoCLIP-B       & 97.19 & 94.12 & 93.86 & \textbf{97.95} & 97.44
                   & 86.55 & 87.81 & 71.26 & \textbf{89.75} & 89.33 \\
\bottomrule
\end{tabular}%
}
\end{table*}

\section{Hierarchy of Hyperbolic Intrinsic Primitives}
\label{app:hierarchy_hyperbolic}

Hyperbolic embeddings encode hierarchy through radial distance from the origin, where larger radii correspond to more specific concepts. Driven by the strong group robustness results, and to further deepen the analysis of the properties of the intrinsic primitives, we examine the distributions of the radius of the representations of attributes, objects, and attribute-object compositions on the manifold of hyperbolic models.

\begin{figure*}[t]
\centering
    \resizebox{\linewidth}{!}{
        \input{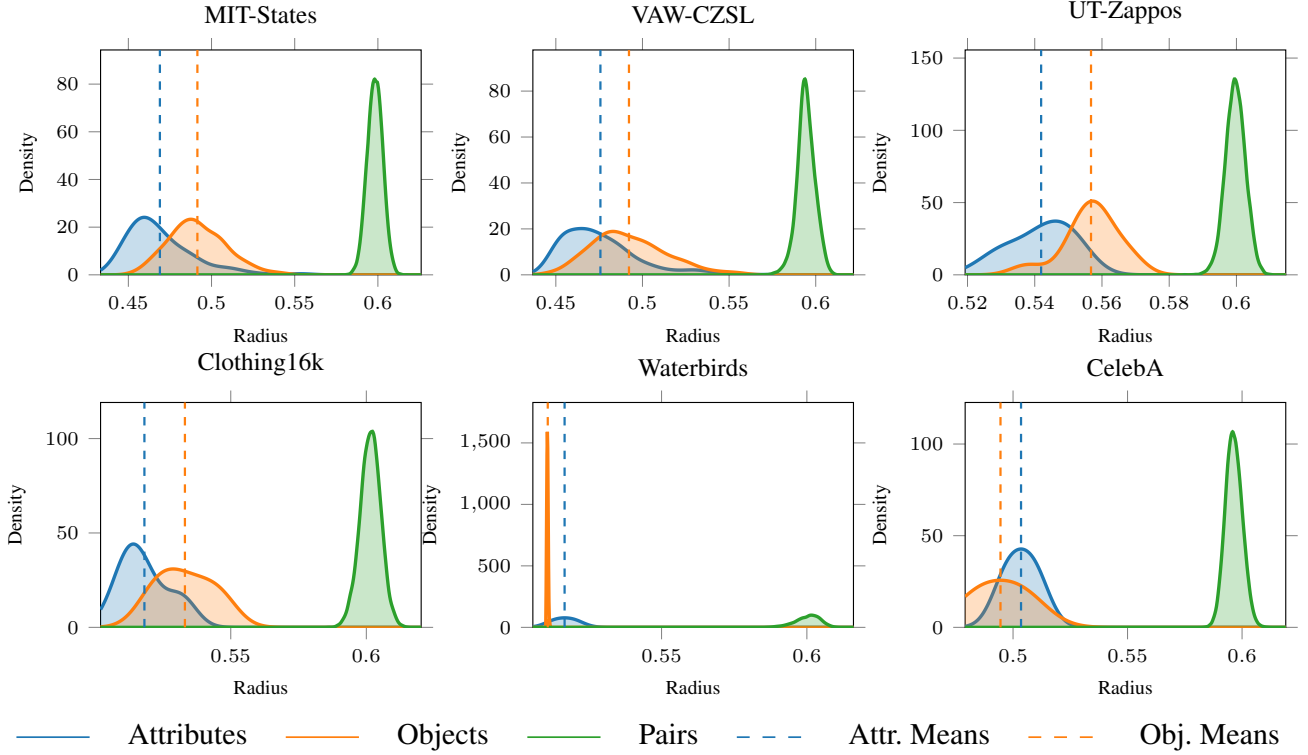}
    }

    \resizebox{\linewidth}{!}{
    \begin{tikzpicture}
    \begin{customlegend}[
        legend columns=-1,
        legend style={draw=none,column sep=1.5ex,font=\scriptsize},
        legend entries={Attributes, Objects, Pairs, Attr.\ Means, Obj.\ Means}
    ]
    \addlegendimage{no markers,attrcol}
    \addlegendimage{no markers,objcol}
    \addlegendimage{no markers,paircol}
    \addlegendimage{no markers,dashed,attrcol}
    \addlegendimage{no markers,dashed,objcol}
    \end{customlegend}
    \end{tikzpicture}
    }
    \caption{
    Radius distributions for HyCoCLIP-B showing attribute primitives, object primitives, and attribute-object compositions in hyperbolic space.
    Dashed lines indicate mean radii. We find, objects intrinsic primitives are consistently occupy the higher radius stratum compared to attributes, except in the case of CelebA and Waterbirds. The pair embeddings themselves are significantly further away, which is consistent with the hierarchical entailment objective of these models.
    }
\label{fig:radius-grit_hycoclip_base}
\end{figure*}

\begin{figure*}[t]
    \centering
    \resizebox{\linewidth}{!}{
        \input{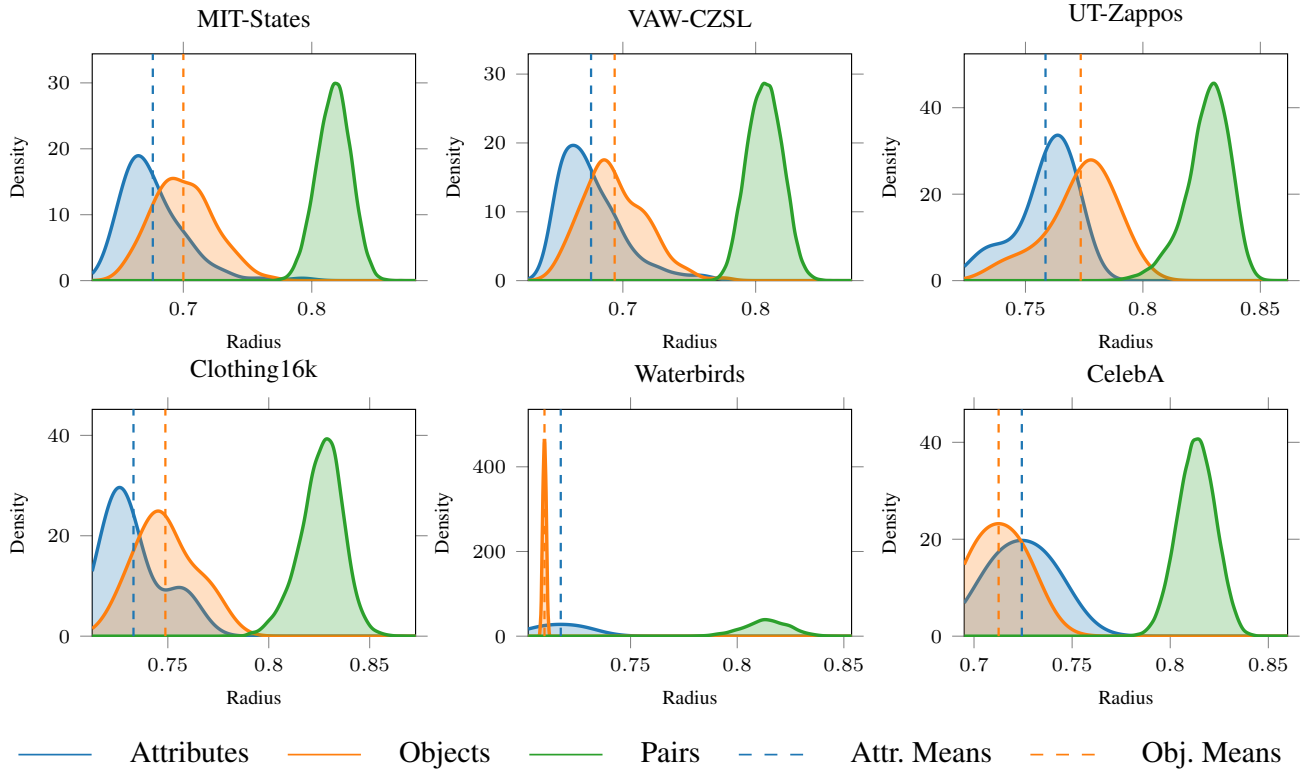}
    }

    \vspace{0.3em}

    \resizebox{\linewidth}{!}{
    \begin{tikzpicture}
    \begin{customlegend}[
        legend columns=-1,
        legend style={
            draw=none,
            column sep=1.5ex,
            font=\scriptsize
        },
        legend entries={
            Attributes,
            Objects,
            Pairs,
            Attr.\ Means,
            Obj.\ Means
        }
    ]
    \addlegendimage{no markers,attrcol}
    \addlegendimage{no markers,objcol}
    \addlegendimage{no markers,paircol}
    \addlegendimage{no markers,dashed,attrcol}
    \addlegendimage{no markers,dashed,objcol}
    \end{customlegend}
    \end{tikzpicture}
    }

    \caption{
    Radius distributions for MERU showing attribute primitives, object primitives, and attribute-object compositions in hyperbolic space.
    Dashed lines indicate mean radii. Similar to HycoCLIP, we find, objects intrinsic primitives are consistently occupy the higher radius stratum compared to attributes, except in the case of CelebA and Waterbirds. The pair embeddings themselves are significantly further away, which is consistent with the hierarchical entailment objective of these models.
    }
    \label{fig:radius-grit_meru_base}
\end{figure*}

We plot the computed distributions for HyCoCLIP in Figure~\ref{fig:radius-grit_hycoclip_base} and for MERU in Figure~\ref{fig:radius-grit_meru_base} for the considered datasets. The embeddings of attribute-object compositions consistently occupy larger radii than their constituent primitives, consistently with the expected semantic hierarchy in hyperbolic spaces: compositions such as {\it blue car} are more specific than either primitives alone.

For MIT-States, VAW-CZSL, UT-Zappos, and Clothing16K, attribute primitives tend to lie closer to the origin and thus represent more general concepts, shared across several objects. Object primitives instead occupy deeper regions of the hierarchy, configuring as slightly more specific and complex semantic concepts in the considered datasets. In contrast, Waterbirds and CelebA data shows the opposite trend, with attributes occupying larger radii than objects. This suggests that attributes (e.g., backgrounds in Waterbirds or facial characteristics in CelebA) add complexity and further specifications to the object concepts in the dataset, possibly inducing biases and dominating the semantic structure of the data. The results of this analysis align with those achieved for group robustness tests where intrinsic primitives show exceptional de-biasing performance for hyperbolic models: the use of object primitives allows to disentangle the complexity induced by the attributes that cause bias and spurious correlations arising in the zero-shot predictions.

From a general perspective, these observations support the effectiveness of intrinsic primitive inference. 
Primitive embeddings occupy denser and better-covered regions of the representation space, whereas compositions correspond to deeper and sparser concepts. By decomposing representations into primitives rather than relying on locally estimated GDE primitives, intrinsic primitive means remain closer to the training distribution and better capture the dominant semantic structure of the dataset. This leads to more stable compositional representations and improved robustness under distribution shifts.

\section{Example Images for Datasets}
\label{app:example_grid}

\begin{figure*}[t]
\centering
\vspace{2mm}
\noindent
\rotatebox[origin=c]{90}{{CelebA}}\hspace{2mm}
\begin{minipage}{0.16\textwidth}
\centering
{\textrm{female dark}}\\[2pt]
\includegraphics[width=\linewidth,height=\linewidth]{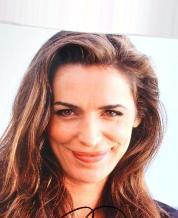}
\end{minipage}
\hspace{1mm}
\begin{minipage}{0.16\textwidth}
\centering
{\textrm{female blonde}}\\[2pt]
\includegraphics[width=\linewidth,height=\linewidth]{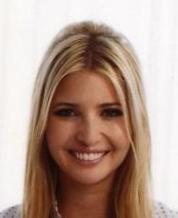}
\end{minipage}
\hspace{1mm}
\begin{minipage}{0.16\textwidth}
\centering
{\textrm{female blonde}}\\[2pt]
\includegraphics[width=\linewidth,height=\linewidth]{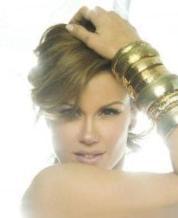}
\end{minipage}
\hspace{1mm}
\begin{minipage}{0.16\textwidth}
\centering
{\textrm{male dark}}\\[2pt]
\includegraphics[width=\linewidth,height=\linewidth]{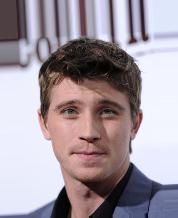}
\end{minipage}
\hspace{1mm}
\begin{minipage}{0.16\textwidth}
\centering
{\textrm{female dark}}\\[2pt]
\includegraphics[width=\linewidth,height=\linewidth]{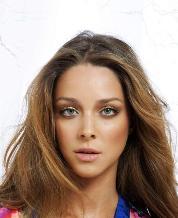}
\end{minipage}
\hspace{1mm}
\par
\vspace{2mm}
\noindent
\rotatebox[origin=c]{90}{{Clothing16k}}\hspace{2mm}
\begin{minipage}{0.16\textwidth}
\centering
{\textrm{green shirt}}\\[2pt]
\includegraphics[width=\linewidth,height=\linewidth]{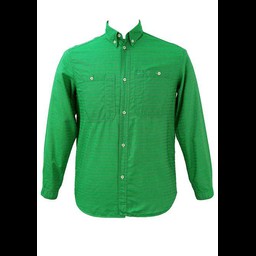}
\end{minipage}
\hspace{1mm}
\begin{minipage}{0.16\textwidth}
\centering
{\textrm{brown shoes}}\\[2pt]
\includegraphics[width=\linewidth,height=\linewidth]{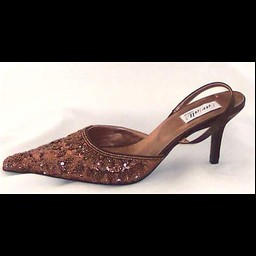}
\end{minipage}
\hspace{1mm}
\begin{minipage}{0.16\textwidth}
\centering
{\textrm{pink skirt}}\\[2pt]
\includegraphics[width=\linewidth,height=\linewidth]{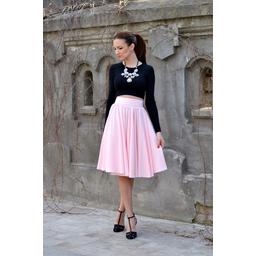}
\end{minipage}
\hspace{1mm}
\begin{minipage}{0.16\textwidth}
\centering
{\textrm{brown shoes}}\\[2pt]
\includegraphics[width=\linewidth,height=\linewidth]{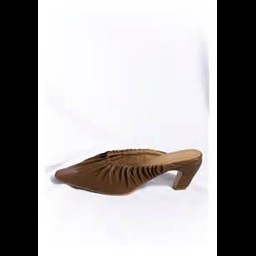}
\end{minipage}
\hspace{1mm}
\begin{minipage}{0.16\textwidth}
\centering
{\textrm{black shirt}}\\[2pt]
\includegraphics[width=\linewidth,height=\linewidth]{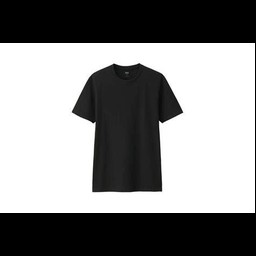}
\end{minipage}
\hspace{1mm}
\par
\vspace{2mm}
\noindent
\rotatebox[origin=c]{90}{{MIT-States}}\hspace{2mm}
\begin{minipage}{0.16\textwidth}
\centering
{\textrm{dented metal}}\\[2pt]
\includegraphics[width=\linewidth,height=\linewidth]{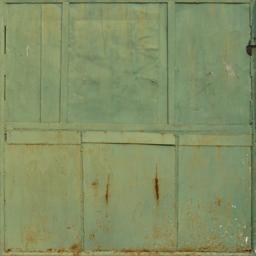}
\end{minipage}
\hspace{1mm}
\begin{minipage}{0.16\textwidth}
\centering
{\textrm{folded bag}}\\[2pt]
\includegraphics[width=\linewidth,height=\linewidth]{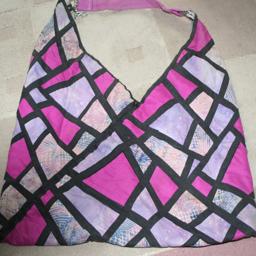}
\end{minipage}
\hspace{1mm}
\begin{minipage}{0.16\textwidth}
\centering
{\textrm{pressed glass}}\\[2pt]
\includegraphics[width=\linewidth,height=\linewidth]{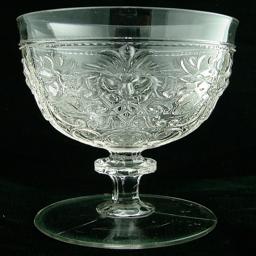}
\end{minipage}
\hspace{1mm}
\begin{minipage}{0.16\textwidth}
\centering
{\textrm{barren creek}}\\[2pt]
\includegraphics[width=\linewidth,height=\linewidth]{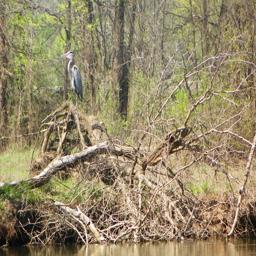}
\end{minipage}
\hspace{1mm}
\begin{minipage}{0.16\textwidth}
\centering
{\textrm{dry well}}\\[2pt]
\includegraphics[width=\linewidth,height=\linewidth]{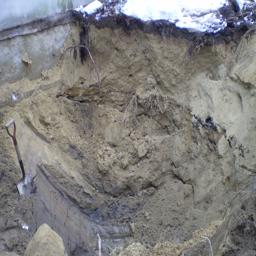}
\end{minipage}
\hspace{1mm}
\par
\vspace{2mm}
\noindent
\rotatebox[origin=c]{90}{{VAW-CZSL}}\hspace{2mm}
\begin{minipage}{0.16\textwidth}
\centering
{\textrm{baby sheep}}\\[2pt]
\includegraphics[width=\linewidth,height=\linewidth]{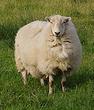}
\end{minipage}
\hspace{1mm}
\begin{minipage}{0.16\textwidth}
\centering
{\textrm{foamy water}}\\[2pt]
\includegraphics[width=\linewidth,height=\linewidth]{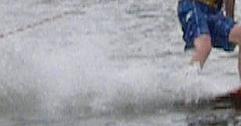}
\end{minipage}
\hspace{1mm}
\begin{minipage}{0.16\textwidth}
\centering
{\textrm{lying lemon}}\\[2pt]
\includegraphics[width=\linewidth,height=\linewidth]{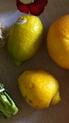}
\end{minipage}
\hspace{1mm}
\begin{minipage}{0.16\textwidth}
\centering
{\textrm{shiny fruit}}\\[2pt]
\includegraphics[width=\linewidth,height=\linewidth]{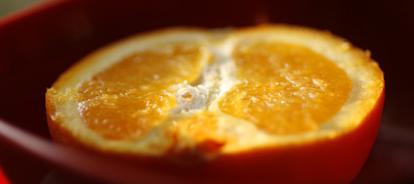}
\end{minipage}
\hspace{1mm}
\begin{minipage}{0.16\textwidth}
\centering
{\textrm{upsidedown cat}}\\[2pt]
\includegraphics[width=\linewidth,height=\linewidth]{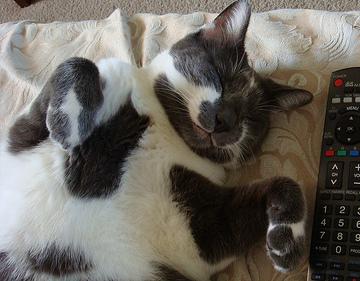}
\end{minipage}
\hspace{1mm}
\par
\vspace{2mm}
\noindent
\rotatebox[origin=c]{90}{{UT-Zapos}}\hspace{2mm}
\begin{minipage}{0.16\textwidth}
\centering
{\textrm{Patent.Leather Shoes.Heels}}\\[2pt]
\includegraphics[width=\linewidth,height=\linewidth]{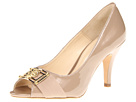}
\end{minipage}
\hspace{1mm}
\begin{minipage}{0.16\textwidth}
\centering
{\textrm{Leather Boots.MidCalf}}\\[2pt]
\includegraphics[width=\linewidth,height=\linewidth]{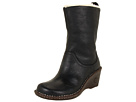}
\end{minipage}
\hspace{1mm}
\begin{minipage}{0.16\textwidth}
\centering
{\textrm{Canvas Shoes.Loafers}}\\[2pt]
\includegraphics[width=\linewidth,height=\linewidth]{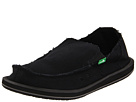}
\end{minipage}
\hspace{1mm}
\begin{minipage}{0.16\textwidth}
\centering
{\textrm{Leather Boots.Ankle}}\\[2pt]
\includegraphics[width=\linewidth,height=\linewidth]{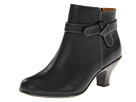}
\end{minipage}
\hspace{1mm}
\begin{minipage}{0.16\textwidth}
\centering
{\textrm{Hair.Calf Shoes.Heels}}\\[2pt]
\includegraphics[width=\linewidth,height=\linewidth]{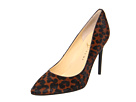}
\end{minipage}
\hspace{1mm}
\par
\vspace{2mm}
\noindent
\rotatebox[origin=c]{90}{{Waterbirds}}\hspace{2mm}
\begin{minipage}{0.16\textwidth}
\centering
{\textrm{water wterbird}}\\[2pt]
\includegraphics[width=\linewidth,height=\linewidth]{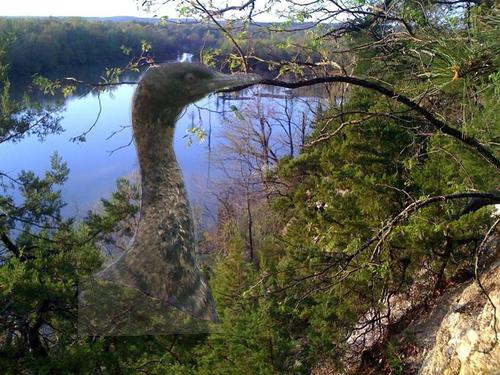}
\end{minipage}
\hspace{1mm}
\begin{minipage}{0.16\textwidth}
\centering
{\textrm{water landbird}}\\[2pt]
\includegraphics[width=\linewidth,height=\linewidth]{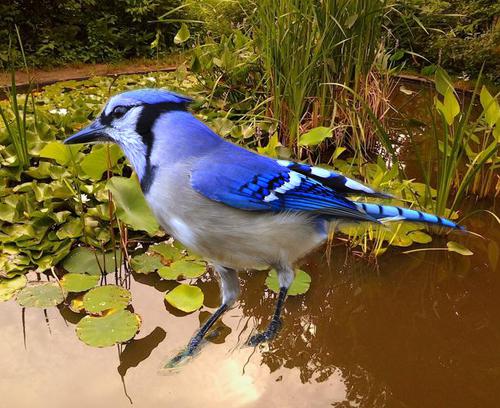}
\end{minipage}
\hspace{1mm}
\begin{minipage}{0.16\textwidth}
\centering
{\textrm{land waterbird}}\\[2pt]
\includegraphics[width=\linewidth,height=\linewidth]{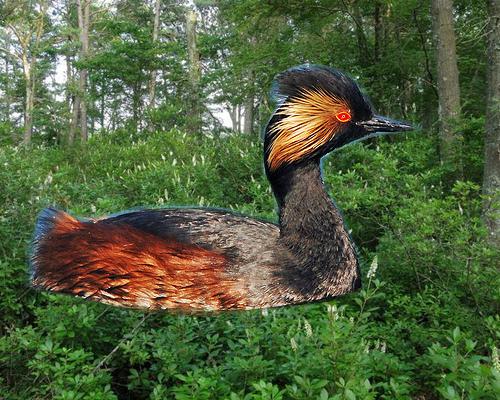}
\end{minipage}
\hspace{1mm}
\begin{minipage}{0.16\textwidth}
\centering
{\textrm{land landbird}}\\[2pt]
\includegraphics[width=\linewidth,height=\linewidth]{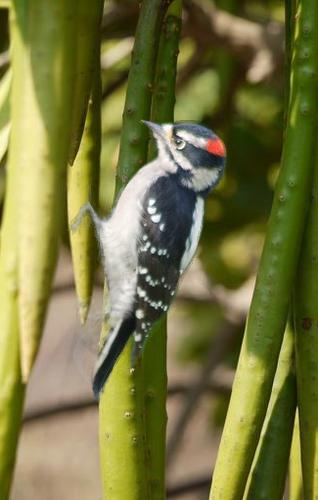}
\end{minipage}
\hspace{1mm}
\begin{minipage}{0.16\textwidth}
\centering
{\textrm{water landbird}}\\[2pt]
\includegraphics[width=\linewidth,height=\linewidth]{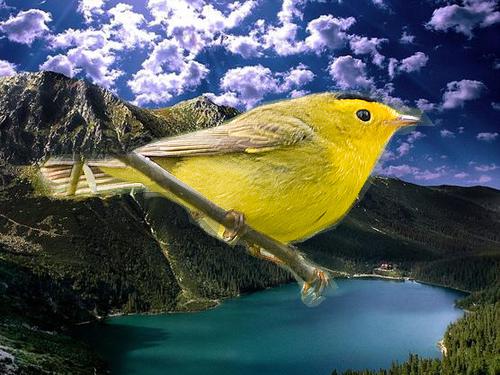}
\end{minipage}
\hspace{1mm}
\par
\caption{Random compositional examples across datasets.}
\label{fig:example_grid}
\end{figure*}

We provide qualitative examples from each dataset used in our compositional zero-shot learning experiments in Figure~\ref{fig:example_grid}. These samples are drawn from the test splits and illustrate the diversity of attribute-object compositions present in each domain.

Overall, most datasets exhibit substantial variation in both attributes and objects, reflecting the combinatorial nature of the task. In particular, Clothing16K contains a broad range of clothing categories and attributes, including different garment types, textures, and visual appearances, as well as a wider variety of poses and viewpoints. This results in higher visual diversity and more challenging compositional generalization settings.

In contrast, UT-Zappos is considerably more constrained, consisting exclusively of footwear categories. Moreover, the visual presentation of objects in this dataset tends to be relatively consistent, with limited variation in pose and viewpoint. This reduced geometric and semantic diversity aligns with our observation of lower compositional variability and helps explain its comparatively different behavior in our results.

\end{document}